\documentclass{article}

\PassOptionsToPackage{numbers, compress}{natbib}
\usepackage[eandd, final]{neurips_2026}

\usepackage[utf8]{inputenc} 
\usepackage[T1]{fontenc}    
\usepackage{amsfonts}       
\usepackage{nicefrac}       
\usepackage{xcolor}         

\usepackage{microtype}
\usepackage{graphicx}
\graphicspath{{figures/}}
\usepackage{subcaption}
\usepackage{booktabs}       
\usepackage{adjustbox} 
\usepackage{wrapfig}
\usepackage{hyperref}
\usepackage{float}
\usepackage{placeins}

\usepackage{tcolorbox}
\tcbuselibrary{breakable, skins} 
\usepackage{capt-of} 

\usepackage{amsmath}
\usepackage{amssymb}
\usepackage{mathtools}
\usepackage{amsthm}
\usepackage{multirow}
\usepackage[capitalize,noabbrev]{cleveref}

\theoremstyle{plain}

\theoremstyle{definition}

\theoremstyle{remark}

\usepackage[textsize=tiny]{todonotes}

\title{OpenVTON-Bench: A Large-Scale High-Resolution \\ Benchmark for Controllable Virtual Try-On Evaluation}

\author{
  \textbf{Jin Li}\textsuperscript{1, 2}\thanks{Equal contribution.} \quad
  \textbf{Tao Chen}\textsuperscript{1}\footnotemark[1] \quad
  \textbf{Kai Wen}\textsuperscript{1, 2} \quad
  \textbf{Siqi Yin}\textsuperscript{1, 2} \\
  \textbf{Shuai Jiang}\textsuperscript{1} \quad
  \textbf{Weijie Wang}\textsuperscript{1} \quad
  \textbf{Jingwen Luo}\textsuperscript{1} \quad
  \textbf{Chenhui Wu}\textsuperscript{1}\thanks{Project lead and corresponding author. \texttt{elbert@renxing.co}} \\
  \textnormal{\textsuperscript{1}Renxing Intelligence, Hangzhou, China} \\
  \textnormal{\textsuperscript{2}Hangzhou Dianzi University, Hangzhou, China}
}

\begin{document}

\maketitle

\begin{abstract}
Recent advances in diffusion models have significantly elevated the visual fidelity of Virtual Try-On (VTON) systems, yet reliable evaluation remains a persistent bottleneck. Traditional metrics struggle to quantify fine-grained texture details and semantic consistency, while existing datasets fail to meet commercial standards in scale and diversity. We present OpenVTON-Bench, a large-scale benchmark comprising approximately 100K high-resolution image pairs (up to $1536\times1536$). The dataset combines Gemini-powered dense captioning with stratified sampling across 20 visual clusters obtained by DINOv3 K-Means. To support reliable evaluation, we propose a multi-modal protocol that measures VTON quality along five interpretable dimensions: background consistency, identity fidelity, texture fidelity, shape plausibility, and overall realism. The protocol integrates VLM-based semantic reasoning with a novel Multi-Scale Representation Metric based on SAM3 segmentation and morphological erosion, enabling the separation of boundary alignment errors from internal texture artifacts. Under the specified evaluation rubric, our representation metric shows strong agreement with human model rankings (Kendall's $\tau$ of 0.833 vs. 0.611 for SSIM). Code and dataset are available at \url{https://github.com/RenxingIntelligence/OpenVTON-Bench}.
\end{abstract}

\section{Introduction}
\label{sec:intro}

Image-based Virtual Try-On (VTON) aims to synthesize photorealistic images of a person wearing a target garment while preserving their identity and the garment's visual attributes~\cite{yang2025omnivton,wang2025mv,choi2021viton}. Driven by the advent of latent diffusion models, modern VTON systems have achieved unprecedented visual quality~\cite{stableviton2024, yang2024texture, PromptDresser2025, SDVTON2024}. Consequently, the research focus is rapidly pivoting from the foundational challenge of \emph{how to generate} to the more critical question of \emph{how to robustly evaluate}. Currently, a perilous gap exists between advanced generative capabilities and outdated evaluation standards, misguiding the field toward optimizing for models that fail in real-world applications.

\begin{figure}[t]
    \centering
    \includegraphics[width=\columnwidth]{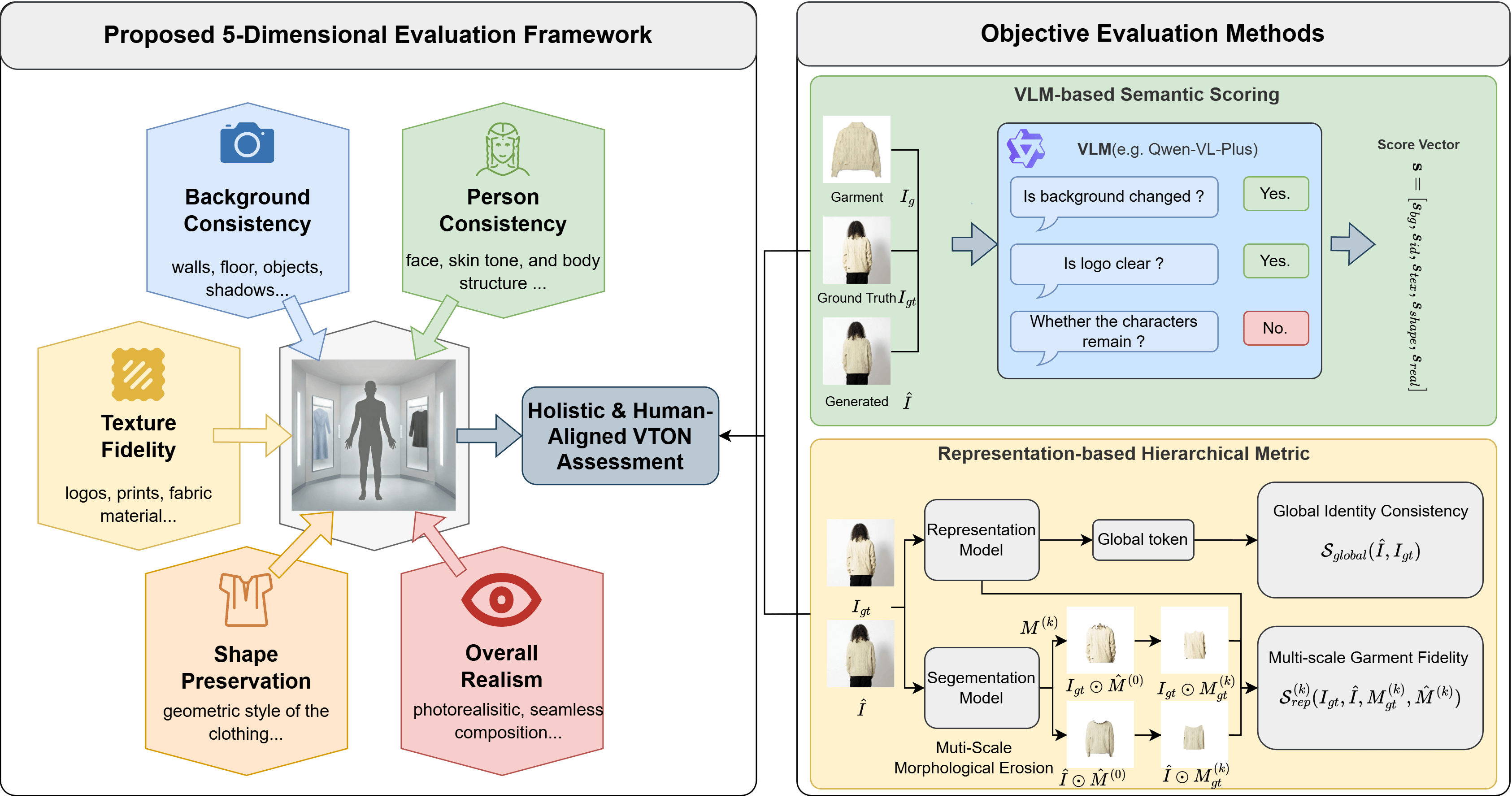} 
    \caption{\textbf{The Proposed Hybrid Evaluation Framework.}
    We move beyond single-scalar metrics by decomposing VTON quality into five human-aligned dimensions. Uniquely, our framework combines \textbf{(Top)} a \textbf{VLM-as-a-Judge} module for semantic auditing with \textbf{(Bottom)} a \textbf{Multi-Scale Representation Metric} that verifies structure-aware garment consistency.
    This synergy ensures both semantic plausibility and accurate garment replication.}
    \label{fig:evaluation_framework}
\end{figure}

This evaluation bottleneck primarily stems from legacy datasets like VITON-HD~\cite{choi2021viton} and DressCode~\cite{DressCode}. While foundational, they suffer from a \textbf{``studio-centric bias''}—featuring clean backgrounds, standardized poses, and restricted resolutions—failing to stress-test models against real-world complexities such as severe occlusions and diverse lighting. Although recent efforts attempt higher resolutions~\cite{wei2024vtonqa}, they remain constrained by small scales or closed-source licenses. To bridge this gap, we introduce \textbf{OpenVTON-Bench}, a large-scale, high-resolution ($1.5$K), and publicly available research benchmark designed for commercial-grade VTON assessment.

Furthermore, conventional metrics like FID~\cite{FID}, SSIM~\cite{SSIM}, and LPIPS~\cite{LPIPS} have become decoupled from human perception. Operating primarily on global statistics or low-level patches, they suffer from \textbf{``semantic blindness''}. For example, a generated image might achieve an excellent FID yet catastrophically distort a brand logo or alter the user's body shape. Such localized artifacts are statistically subtle to a CNN but immediately disqualifying in commercial applications.

To address these limitations, we propose a \textbf{Multi-Modal Evaluation Protocol} that unites high-level semantic reasoning with low-level structural verification. First, we adapt the \textbf{VLM-as-a-Judge} approach to reference-based VTON evaluation. State-of-the-art Vision-Language Models (VLMs)~\cite{bai2025qwen2, zhu2025internvl3} can follow structured, fine-grained multimodal evaluation instructions to scrutinize images and identify semantic misalignments. However, because VLMs often lack spatial precision for subtle textile deformations~\cite{VTBench2025}, we complement them with a \textbf{Multi-Scale Representation Metric}. By applying morphological erosion to SAM3-generated masks~\cite{sam3}, this metric constructs a hierarchy of interior regions to explicitly decouple boundary alignment errors from internal texture artifacts.

By operationalizing this hybrid protocol, we systematically decompose VTON quality into five orthogonal axes (Figure~\ref{fig:evaluation_framework}): \textit{background consistency}, \textit{identity fidelity}, \textit{texture fidelity}, \textit{shape plausibility}, and \textit{overall realism}. Together, these dimensions enable a granular and interpretable diagnosis of VTON deficiencies that transcends coarse holistic scoring. Our contributions are summarized as follows:

\begin{itemize}
    \setlength{\itemsep}{1pt}
    \setlength{\parskip}{0pt}
    \setlength{\parsep}{0pt}
    
    \item \textbf{OpenVTON-Bench:} We release a high-fidelity benchmark of $\sim$100K high-resolution ($1536^2$) image pairs, processed into \textbf{standardized triplets} with dense captions and stratified sampling across 20 DINOv3 K-Means visual clusters.
    \vspace{10pt}
    \item \textbf{Hybrid Evaluation Paradigm:} We integrate VLM-as-a-Judge for semantic reasoning with a Multi-Scale Representation Metric for structural verification, validated against human model rankings under a shared VTON rubric.
    \vspace{10pt}
    \item \textbf{Diagnostic Analysis:} Benchmarking state-of-the-art models reveals that while modern diffusion models excel in photorealism, they frequently hallucinate fine-grained textures—a critical insight revealed only through our dual-track evaluation.
\end{itemize}

\FloatBarrier
\begin{figure}[!t]
    \centering
    \includegraphics[width=0.95\textwidth]{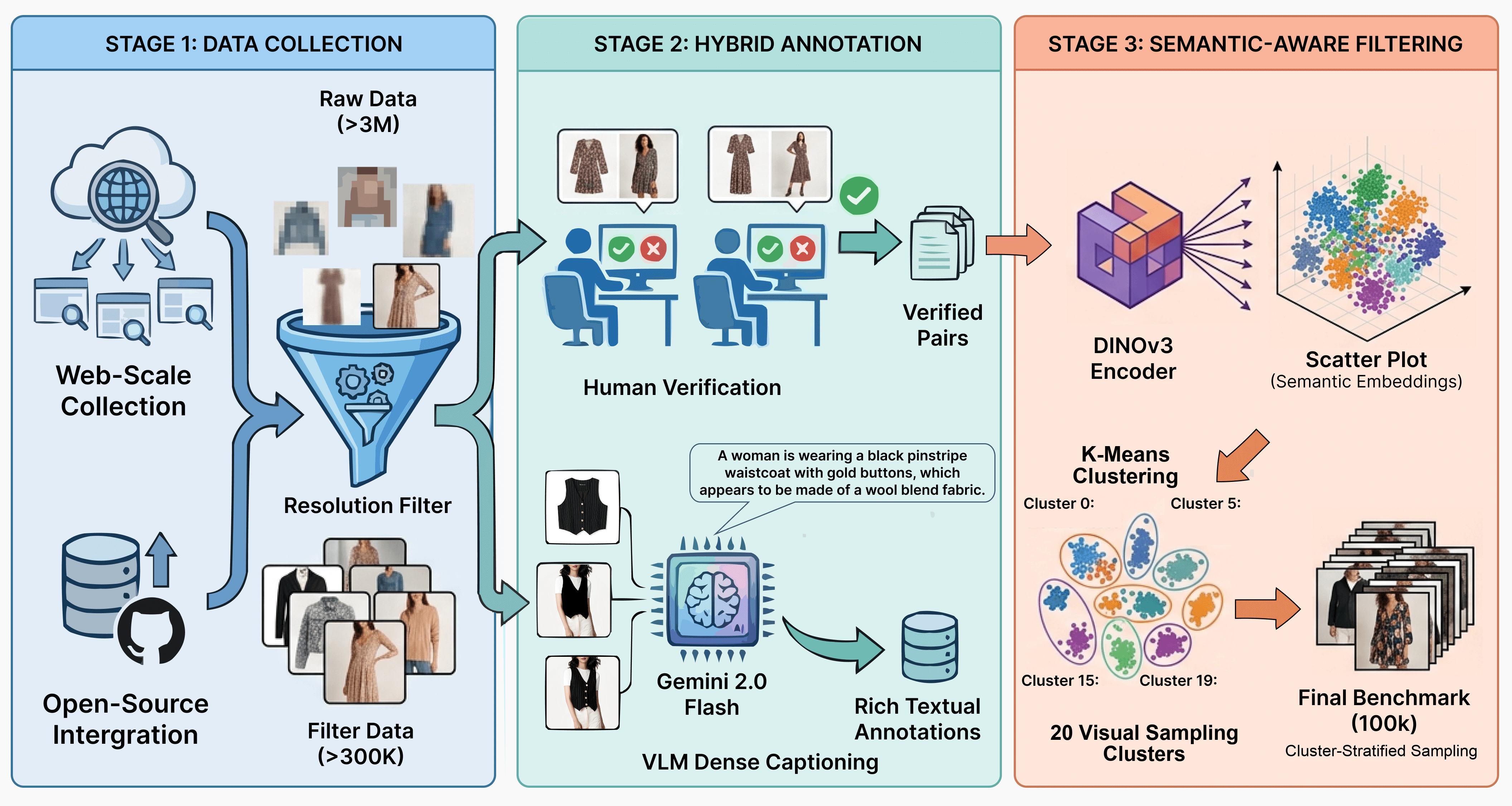} 
    \caption{\textbf{Data Construction Pipeline of OpenVTON-Bench.} The process consists of three stages: (1) Large-scale raw data aggregation from diverse sources; (2) Hybrid annotation combining human verification for pair alignment and VLM-based dense captioning; (3) Stratified sampling across 20 DINOv3 K-Means visual clusters, subject to source availability.}
    \label{fig:data_pipeline}
\end{figure}
\FloatBarrier
\section{Related Works}
\label{sec:related}
\subsection{Virtual Try-On: Methods and Benchmarks}
The evolution of Virtual Try-On (VTON) has transitioned from warping-based synthesis to generative modeling. 
Early GAN-based methods~\cite{VITON,CP-VTON,VITON-GAN,TryOnGAN,choi2021viton,issenhuth2020not,ge2021disentangled,HR-VITON} relied on explicit cloth warping~\cite{TPS,STN,flownet}, which often fails to handle complex poses or high resolutions due to geometric limitations. 
Conversely, recent diffusion-based approaches~\cite{LaDIVTON2023,TryOnDiffusion2023,zheng2024viton,SDVTON2024,stableviton2024,catdm2024,catvton2024,dsvton2025,catv2ton2025,PromptDresser2025} formulate VTON as conditional inpainting, significantly improving photorealism and non-rigid deformation handling. 
However, despite these architectural gains, a critical \emph{data bottleneck} persists: models struggle with misalignment at resolutions beyond $1$K due to the scarcity of high-quality training data.

Existing benchmarks struggle to balance supervision quality with environmental diversity. 
Controlled paired datasets~\cite{MPV,choi2021viton,DressCode} offer reliable ground truth but lack pose and background variation, whereas in-the-wild collections~\cite{deepfashion,UPT,ESF,StreetTryon,SHHQ,LH-400K} provide realism but lack paired supervision, complicating faithful evaluation.
Although recent efforts like VTONQA~\cite{wei2024vtonqa} and VTBench~\cite{VTBench2025} explore higher resolutions and refined metrics, they remain constrained by limited scale, instability from model-generated hallucinations, or closed-source policies. 
To address these limitations, we introduce OpenVTON-Bench, a large-scale ($\sim 100$K), high-resolution ($1.5$K) benchmark that combines paired supervision with in-the-wild diversity, enabling rigorous evaluation of next-generation systems.

\subsection{Evaluation Protocols: From Pixels to Semantics}
Evaluating virtual try-on requires preserving garment fidelity while integrating it realistically with the person. Conventional metrics include PSNR, SSIM, LPIPS, and FID~\cite{PSNR,SSIM,LPIPS,FID}. Pixel-wise scores penalize legitimate spatial variations, while distributional metrics can overlook local texture errors. CLIP-based scores~\cite{song2024imagebasedvirtualtryonsurvey} provide semantic alignment but may miss fine garment details. MLLM-as-a-Judge~\cite{chen2024mllm} studies multimodal judging capabilities and biases, and VIEScore~\cite{ku2024viescore} uses MLLMs for explainable evaluation of conditional image synthesis. VTON-VLLM~\cite{wan2025vtonvllm} learns from human annotations to evaluate VTON results and guide synthesis toward human preferences. Other work explores improving VLM judges without human annotations~\cite{lin2025self}.
Building on these approaches, our contribution is a VTON-specific hybrid protocol combining reference-based VLM scoring with SAM3 masks, progressive erosion, and DINOv3 features, validated against human model rankings under a shared rubric.
\section{OpenVTON-Bench}
\label{sec:bench}
Constructing a benchmark that accurately reflects the demands of commercial virtual try-on requires moving beyond simple studio settings. In this section, we detail the construction pipeline of \textbf{OpenVTON-Bench}, covering data acquisition, hybrid annotation, semantic-aware filtering, and detailed statistics. The overall construction pipeline is illustrated in Figure~\ref{fig:data_pipeline}.

\subsection{Data Collection}
We combine two complementary data streams to cover diverse body shapes, poses, clothing styles, and backgrounds, applying common resolution criteria to the collected pairs.

\textbf{Open-Source Integration.} We curated garment--person pairs from the IG Pair Dataset in IMAGDressing-v1~\cite{IMAGDressing-v1}. Manual re-screening retained one-to-one garment--person correspondences and removed redundant poses for the same garment. This stream contributes examples from social media, complementing the e-commerce images collected below.

\textbf{Web-Scale Collection.} We collected images from publicly available e-commerce sources following standard practices in benchmark construction~\cite{LAION}. Aesthetic scoring and face detection screened low-quality images and images without visible faces. These sources broaden the range of clothing styles, lighting conditions, and backgrounds represented in the candidate pool.

The dataset is intended for non-commercial research under its release license. We provide a takedown mechanism for copyright holders to request removal of specific samples. Facial regions underwent generative anonymization to protect identity while preserving the body and garment structure needed for evaluation.

The initial raw collection exceeded \textbf{3,000,000} samples. 
We applied a strict resolution constraint, retaining only images whose \emph{height and width are at least} $1024$ pixels, while \emph{the longer side does not exceed} $1536$ pixels.
This criterion allows both square and rectangular images, and ensures that all retained samples satisfy the high-fidelity requirements of modern commercial VTON applications.
After this filtering stage, approximately 300,000 samples remained.
The final 99,925 image pairs were obtained through a subsequent stratified sampling process (Section~\ref{sec:filtering}).

\subsection{Hybrid Annotation Pipeline}
High-quality VTON requires not only precise image pairs but also rich semantic context. We employed a Human-AI hybrid annotation strategy.

\textbf{Human-in-the-Loop Pair Verification.}~
The core of VTON datasets is the correspondence between the \textit{in-shop garment} (source) and the \textit{person wearing it} (reference). Automated matching often fails with complex layering. We deployed annotators to verify all candidate pairs in the 300K pool, discarding samples with mismatched items, severe occlusions, or missing views. 

\textbf{VLM-Powered Dense Captioning.}~
To support text-guided editing and multimodal evaluation, we utilized Gemini 2.0 Flash~\cite{gemini}—selected for strong performance on fine-grained visual attribute extraction and cost efficiency at scale—to generate comprehensive descriptions for each garment. To enhance the precision of these descriptions, we implemented a coarse-to-fine prompting strategy.

This process begins with a coarse-grained classification prompt that categorizes the primary garment in an image as either an `upper-body' or `lower-body' item. For this initial sort, full-body garments such as dresses are classified as upper-body items that we found to produce more coherent captions. Following this classification, a specialized prompt is conditionally applied. For upper-body garments, the VLM is guided to extract \textit{Structure} (sleeve length, neckline), \textit{Texture} (fabric, patterns), and \textit{Design Details} (logos, ruffles), while for lower-body garments, the focus shifts to analogous details such as \textit{Structure} (fit, cut), \textit{Texture} (denim, wash), and \textit{Design Details} (pockets, embroidery).

This two-tiered methodology ensures that the extracted attributes are contextually relevant, generating rich textual annotations that far exceed the granularity of traditional, monolithic label-based systems.
\subsection{Semantic-Aware Filtering via DINOv3}
\label{sec:filtering}
A challenge in fashion datasets is the long-tail distribution problem—simple items like "white t-shirts" often dominate, while complex textures are underrepresented. To balance sampling across visual groups, we implemented a clustering pipeline utilizing Self-Supervised Learning (SSL).

\textbf{Semantic Embedding Extraction.} ~
We fed all verified garment images into the \textbf{DINOv3 (ViT-H+)}~\cite{dinov3} encoder. DINOv3 was selected for its superior ability to capture holistic semantic structures and object-level features compared to CLIP, which focuses more on text alignment. This yields a robust visual representation invariant to slight deformations.

\textbf{K-Means Clustering \& Stratified Sampling.} ~
We applied K-Means ($K=20$) to the DINOv3 embeddings to obtain \textbf{20 visual sampling clusters} (C0--C19). Stratified sampling across these clusters yielded the final \textbf{99,925} samples, referred to as the \textbf{``100K''} dataset. Sampling is approximately balanced across clusters, subject to source availability.
The released \texttt{category} field instead contains \textbf{14 coarse labels} obtained by deterministic keyword matching over Gemini captions. These metadata labels differ from the visual clusters used for final sampling. A coarse label can span several clusters, and label frequencies need not be uniform. Implementation details are provided in Appendix~\ref{app:data_balancing_annotation}.

\subsection{Benchmark Overview and Statistics}
\label{sec:stats}
The final OpenVTON-Bench comprises 99,925 high-resolution image pairs, establishing itself as one of the largest VTON benchmarks with consistent high fidelity. As summarized in Table~\ref{tab:dataset_comparison}, OpenVTON-Bench maintains a minimum resolution of $1024 \times 1024$, with images reaching up to $1536 \times 1536$—a critical requirement for evaluating the fine-grained texture generation capabilities of generative models.

Beyond scale, OpenVTON-Bench provides substantially richer annotations than its predecessors. Each sample is accompanied by dense semantic captions totaling over 3 million words, capturing nuanced attributes such as fabric texture, pattern complexity, and design details that are absent from traditional label-based annotations. This enables not only more comprehensive evaluation but also opens avenues for text-guided VTON research.

\begin{wrapfigure}{r}{0.5\textwidth}
    \vspace{-12pt}
    \centering
    \includegraphics[width=\linewidth]{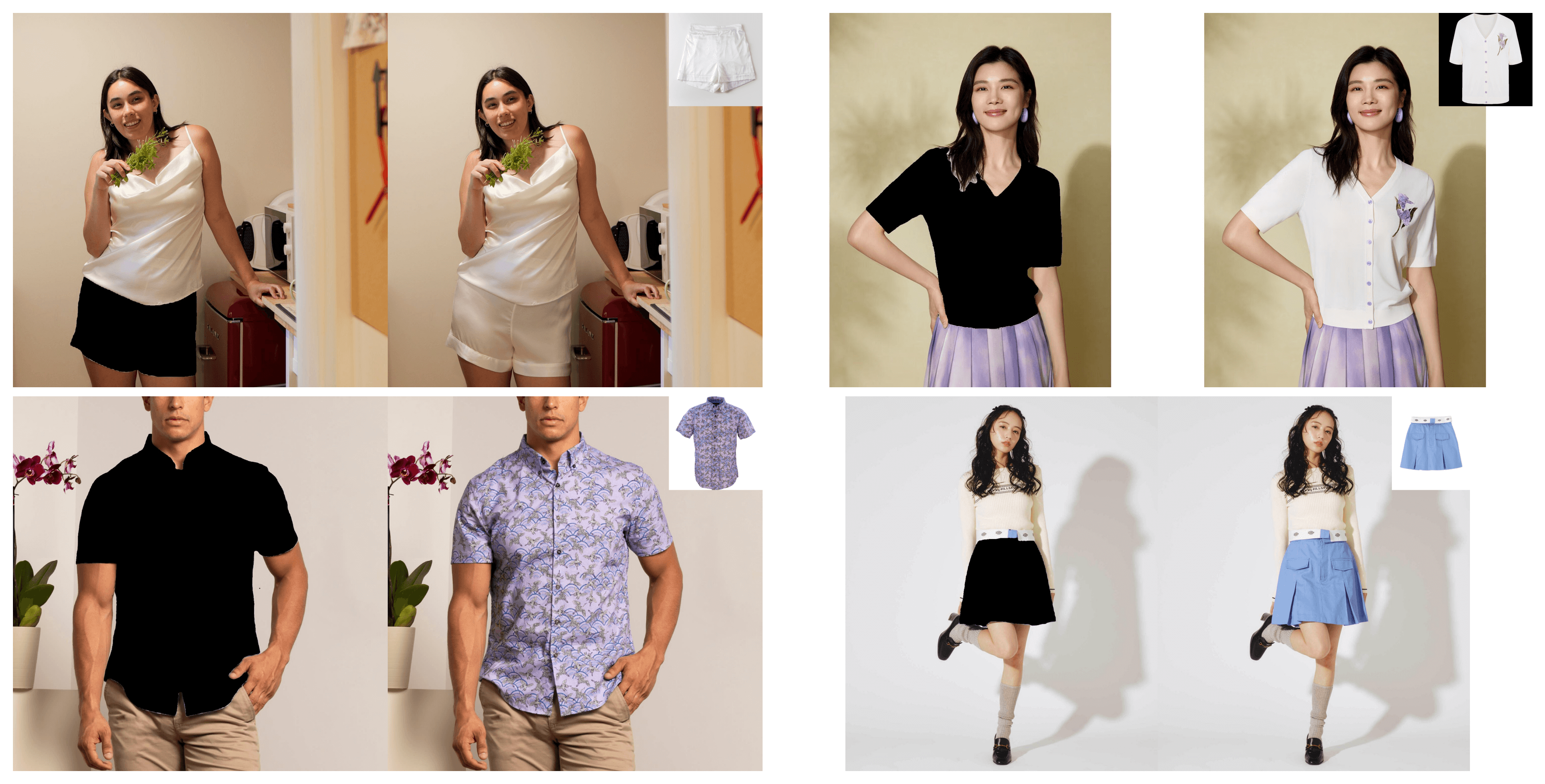}
    
    \vspace{-5pt}
    \small 
    \caption{Representative examples from OpenVTON-Bench.}
    \label{fig:example_samples}
    \vspace{-10pt}
\end{wrapfigure}

The dataset is partitioned by stratifying within the 20 visual sampling clusters, preserving their proportions across the training, validation, and test splits. This partitioning supports evaluation across diverse visual groups without assuming uniform frequencies for the 14 caption-derived labels. Figure~\ref{fig:example_samples} illustrates samples from OpenVTON-Bench, highlighting the diversity of garment types, poses, and visual appearances.
Additional large-scale grid visualizations are provided in Figure~\ref{fig:dataset_samples} of Appendix~\ref{app:data_vis}.
\section{Evaluation Protocol}
\label{sec:eval}

To rigorously assess the quality of virtual try-on synthesis, we propose a multi-modal evaluation protocol that goes beyond traditional pixel-wise comparisons. Our protocol integrates semantic reasoning from VLMs, hierarchical feature analysis from self-supervised learning representations, and conventional statistical metrics, enabling a comprehensive and fine-grained evaluation.

\subsection{Preliminaries and Notation}
Let $\mathcal{D} = \{(I_p, I_g, I_{gt})\}$ denote the evaluation dataset, where $I_p$ is the cloth-agnostic person image, $I_g$ is the target garment image, and $I_{gt}$ is the ground-truth try-on result. A virtual try-on model $G$ produces a synthesized image as defined in Eq.~\ref{eq:vton_generation}.
\begin{equation}
\hat{I} = G(I_p, I_g),
\label{eq:vton_generation}
\end{equation}
The generated image $\hat{I}$ is evaluated against $I_{gt}$. We denote by $\Phi(\cdot)$ a frozen SSL image encoder and by $\mathcal{SAM}(\cdot)$ a frozen segmentation foundation model used for garment localization. The overall evaluation space is defined as $\mathcal{E}=\{\mathcal{E}_{\text{VLM}}, \mathcal{E}_{\text{Rep}}, \mathcal{E}_{\text{Pix}}\}$, corresponding to semantic-, representation-, and pixel-level metrics, respectively.

\subsection{Objective Evaluation}

\subsubsection{VLM-based Semantic Scoring}
\label{sec:metric_vlm}
Human perception of realism primarily depends on high-level semantic consistency. To capture this, we employ a Vision-Language Model (Qwen-VL-Plus~\cite{Qwen-VL}) as a surrogate perceptual judge. For each test case, the model evaluates a visual triplet $(I_g, I_{gt}, \hat{I})$ —where $\hat{I}$ replaces the masked input—together with a task-specific prompt $\mathcal{T}$ and outputs a five-dimensional semantic score vector as formulated in Eq.~\ref{eq:vlm_score}.
\begin{equation}
\mathbf{s} =[s_{bg}, s_{id}, s_{tex}, s_{shape}, s_{real}] 
= \mathcal{V}(I_g, I_{gt}, \hat{I}; \mathcal{T}),
\label{eq:vlm_score}
\end{equation}
 Each scalar score lies in $[1,5]$ and corresponds to background consistency, identity preservation, texture fidelity, shape preservation, and overall realism, respectively. This formulation allows the evaluation of semantic attributes (e.g., logo clarity or garment-category correctness) that are not accessible to conventional CNN-based metrics.

\subsubsection{Representation-based Metrics}
\label{sec:metric_rep}
Pixel-level distances are overly sensitive to minor spatial misalignments that are perceptually negligible. To robustly assess identity preservation and garment texture fidelity, we introduce representation-based metrics built upon DINOv3~\cite{dinov3} features and mask guidance from SAM3~\cite{sam3}.

\paragraph{Global Identity Consistency.}
Overall visual coherence is evaluated by computing the cosine similarity between global image embeddings extracted by $\Phi(\cdot)$. Specifically, the global identity consistency score is defined as
\begin{equation}
S_{\text{global}}(\hat{I}, I_{gt}) =
\frac{\Phi(\hat{I})^\top \Phi(I_{gt})}
{\|\Phi(\hat{I})\|_2 \, \|\Phi(I_{gt})\|_2},
\label{eq:S_global}
\end{equation}
where Eq.~\ref{eq:S_global} measures the alignment between the generated image and the ground truth in the global feature space. A higher $S_{\text{global}}$ indicates better preservation of person identity and overall structural coherence.

\paragraph{Multi-scale Garment Fidelity.}
\label{sec:garment_fidelity}
To disentangle boundary misalignment from internal texture distortion, we further propose a mask-guided multi-scale garment evaluation. First, a binary garment mask is extracted from the ground-truth image as defined in Eq.~\ref{eq:mask_extraction}.
\begin{equation}
M_{gt} = \mathcal{SAM}(I_{gt}).
\label{eq:mask_extraction}
\end{equation}
Similarly, we apply the same segmentation process to the generated image $\hat{I}$ to obtain its corresponding mask $\hat{M}$.
Based on these masks, we generate sets of $K$ nested masks by progressive morphological erosion. Given a structural element $B$, the $k$-th eroded mask is defined as:
\begin{equation}
M_{*}^{(k)} = M_{*} \ominus \bigl(\underbrace{B \oplus \cdots \oplus B}_{k\ \text{times}}\bigr),
\label{eq:mask_erosion}
\end{equation}
where $M_{*}$ denotes either $M_{gt}$ or $\hat{M}$, and $\ominus$ and $\oplus$ denote erosion and dilation, respectively. As described in Eq.~\ref{eq:mask_erosion}, smaller values of $k$ retain garment boundary regions, while larger values focus on the interior fabric area.

For each scale $k$, we compute a masked feature similarity score in the SSL feature space:
\begin{equation}
S_{\text{rep}}^{(k)} =
\frac{\Phi(\hat{I} \odot \hat{M}^{(k)})^\top \Phi(I_{gt} \odot M_{gt}^{(k)})}
{\|\Phi(\hat{I} \odot \hat{M}^{(k)})\|_2 \, \|\Phi(I_{gt} \odot M_{gt}^{(k)})\|_2},
\label{eq:S_rep_1}
\end{equation}
where $\odot$ denotes element-wise multiplication. According to Eq.~\ref{eq:S_rep_1}, a higher $S_{\text{rep}}^{(k)}$ reflects better garment texture fidelity at the corresponding spatial scale, enabling fine-grained diagnosis of boundary versus interior errors.

\subsubsection{Pixel-based Statistical Metrics}
For completeness and compatibility with prior benchmarks, we additionally report conventional pixel-level and distribution-based metrics, including PSNR, SSIM, LPIPS, and FID~\cite{PSNR,SSIM,LPIPS,FID}. These metrics are treated as auxiliary references, and their limitations in capturing perceptual and semantic fidelity are discussed in Sec.~\ref{sec:intro}.

\begin{figure}[htbp]
    \centering
    \includegraphics[width=\linewidth]{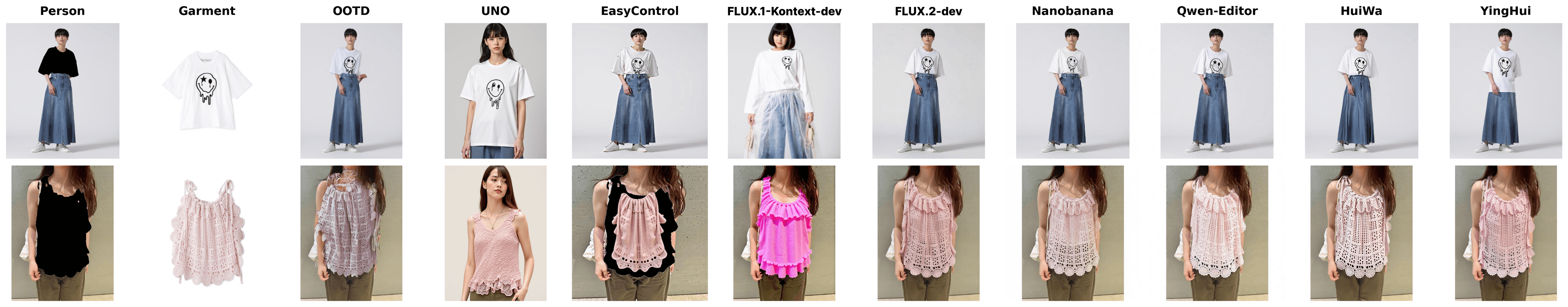}
    \caption{Qualitative comparison of state-of-the-art methods on OpenVTON-Bench.}
    \label{fig:comparison_3131}
\end{figure}

\subsection{Subjective Evaluation}
To validate the proposed metrics, 76 participants provided 92,072 evaluation records for randomly sampled result triplets $(I_g, I_{gt}, \hat{I})$. Participants independently rated each triplet on a five-point Likert scale along the same five dimensions as the VLM (Eq.~\ref{eq:vlm_score}), without seeing method identities or VLM outputs, scores, or rationales. Each triplet received at least two evaluations, which were averaged. We compute Spearman's rank correlation, Kendall's Tau, and Pearson's correlation between model-level human scores and objective metrics. This study assesses model-ranking consistency and score agreement under the specified VTON rubric; it does not measure unconstrained human preference.
\section{Experimental Results}
\label{sec:exp}

In this section, we conduct a comprehensive benchmarking of state-of-the-art virtual try-on systems on OpenVTON-Bench.
We evaluate \textbf{nine} representative methods, covering both diffusion-based paradigms and editing-based frameworks.
Our analysis aims to answer two fundamental questions: (1) How well do current models handle the fine-grained semantic and texture challenges proposed in our dataset? (2) How effectively do our proposed evaluation metrics align with human perception compared to traditional protocols?

\begin{table}[htbp]
\centering
\caption{\textbf{Semantic Evaluation via VLM and Human Annotators.}
We report the scores (scale 1--5) across five semantic dimensions: Background ($s_{bg}$), Identity ($s_{id}$), Texture ($s_{tex}$), Shape ($s_{shape}$), and Overall Realism ($s_{real}$).
\textbf{Bold} indicates the best result, and \underline{underline} indicates the second best.}
\label{tab:vlm_eval}

\begin{adjustbox}{width=0.98\textwidth}
\begin{tabular}{lcccccccccccc}
\toprule
\multirow{2}{*}{\textbf{Method}}
& \multicolumn{2}{c}{\textbf{$s_{bg}$}}
& \multicolumn{2}{c}{\textbf{$s_{id}$}}
& \multicolumn{2}{c}{\textbf{$s_{tex}$}}
& \multicolumn{2}{c}{\textbf{$s_{shape}$}}
& \multicolumn{2}{c}{\textbf{$s_{real}$}}
& \multicolumn{2}{c}{\textbf{$s_{avg}$}} \\
\cmidrule(lr){2-3} \cmidrule(lr){4-5} \cmidrule(lr){6-7} \cmidrule(lr){8-9} \cmidrule(lr){10-11} \cmidrule(lr){12-13}
& \textbf{VLM} & \textbf{Human} & \textbf{VLM} & \textbf{Human} & \textbf{VLM} & \textbf{Human} & \textbf{VLM} & \textbf{Human} & \textbf{VLM} & \textbf{Human} & \textbf{VLM} & \textbf{Human} \\
\midrule
OOTD~\cite{ootdiffusion} & 4.074 & 4.608 & 3.685 & 3.832 & 3.278 & 3.217 & 3.898 & 3.697 & 3.794 & 3.595 & 3.746 & 3.790 \\
UNO~\cite{UNO} & 4.167 & 3.220 & 3.749 & 2.308 & 3.577 & 2.364 & 4.141 & 2.907 & 4.067 & 3.913 & 3.940 & 2.942 \\
EasyControl~\cite{easycontrol} & 4.273 & 4.459 & 3.824 & 3.496 & 3.506 & 3.163 & 4.105 & 3.824 & 4.055 & 3.900 & 3.953 & 3.768 \\
FLUX.1-Kontext-dev~\cite{flux1} & 4.428 & 4.406 & 3.957 & 3.535 & 3.574 & 3.122 & 4.158 & 3.768 & 4.137 & 4.060 & 4.051 & 3.778 \\
FLUX.2-dev~\cite{flux2} & 4.593 & \underline{4.734} & 4.226 & 4.460 & 4.007 & 3.914 & 4.446 & 4.494 & 4.409 & 4.582 & 4.336 & 4.437 \\
NanobananaPro~\cite{nano} & 4.610 & 4.722 & 4.253 & \textbf{4.689} & 4.019 & \underline{4.225} & 4.457 & \underline{4.646} & 4.418 & \textbf{4.694} & 4.351 & \underline{4.595} \\
Qwen-Editor~\cite{qweneditor} & \underline{4.617} & 4.729 & \underline{4.270} & 4.606 & \underline{4.046} & 4.132 & \underline{4.472} & 4.590 & \underline{4.431} & 4.666 & \underline{4.367} & 4.545 \\
HuiWa~\cite{huiwa} & 4.611 & 4.682 & 4.256 & 4.538 & \textbf{4.050} & 4.018 & 4.465 & 4.546 & 4.424 & 4.664 & 4.361 & 4.490 \\
YingHui~\cite{yinghui} & \textbf{4.624} & \textbf{4.735} & \textbf{4.277} & \underline{4.685} & \textbf{4.050} & \textbf{4.252} & \textbf{4.473} & \textbf{4.675} & \textbf{4.437} & \underline{4.693} & \textbf{4.372} & \textbf{4.608} \\
\bottomrule
\end{tabular}
\end{adjustbox}
\end{table}

\subsection{Semantic and Perceptual Evaluation}
\label{sec:exp_vlm}

We first assess the semantic alignment and realism using our VLM-based evaluation protocol and Human Evaluation.
As reported in Table~\ref{tab:vlm_eval}, we make the following observations.

\textbf{VLM Scores Correlate with Human Judgments.}~
Our VLM scores show strong agreement with human model rankings under the shared rubric, although individual scores differ. Notably, \textbf{YingHui} achieves the highest average scores in both VLM ($S_{avg}=4.372$) and human evaluation ($S_{avg}=4.608$). This agreement supports the specified judge and scoring protocol for model comparison on this benchmark.

\textbf{The ``Texture-Realism" Gap.}~
A critical trend is observed in the \textit{Texture} dimension ($S_{tex}$). While general-purpose diffusion models like FLUX.1-Kontext-dev achieve high scores in \textit{Background} ($4.428$) and \textit{Realism} ($4.137$), their performance drops significantly in \textit{Texture} ($3.574$). This indicates that while large-scale pre-training yields photorealism, it struggles with the zero-shot preservation of specific garment patterns.
In contrast, commercial-grade systems (e.g., YingHui, HuiWa), which typically leverage large-scale proprietary try-on data, exhibit a much more balanced performance profile across all dimensions.
This sharp contrast verifies the necessity of highly diverse, domain-specific benchmarks like OpenVTON-Bench to push the boundaries of open-source models.

\subsection{Fine-Grained Texture Fidelity}
\label{sec:exp_rep}

Pixel-based metrics often fail to distinguish between correct texture synthesis and mere boundary alignment. To address this, we employ our Representation-based Similarity ($\mathcal{S}_{rep}$) with progressive mask erosion, which forces the evaluation to focus on the \textit{inner} garment details rather than edge contrast.
The results are presented in Table~\ref{tab:repr_eval}.

\setlength{\columnsep}{15pt} 
\begin{wraptable}{r}{0.52\textwidth}
    \vspace{-13pt} 
    \small 
    \caption{\textbf{Representation-based Similarity Evaluation.}
    $\mathcal{S}_{\text{global}}$ measures holistic semantic consistency.
    $\mathcal{S}_{\text{rep}}^{(k)}$ denotes garment-level similarity under increasing mask erosion levels ($0 \rightarrow 3$), isolating texture quality from boundary artifacts.
    }
    \label{tab:repr_eval}
    
    \resizebox{\linewidth}{!}{
    \begin{tabular}{l|c|ccccc|c}
    \toprule
    \textbf{Method} 
    & $\mathcal{S}_{\text{global}}$ 
    & $\mathcal{S}_{\text{rep}}^{(0)}$
    & $\mathcal{S}_{\text{rep}}^{(1)}$ 
    & $\mathcal{S}_{\text{rep}}^{(2)}$ 
    & $\mathcal{S}_{\text{rep}}^{(3)}$ 
    & \textbf{$\bar{\mathcal{S}}_{rep}$}
    & \textbf{$\bar{\mathcal{S}}$}\\
    \midrule
    OOTD~\cite{ootdiffusion} & 0.844 & 0.797 & 0.755 & 0.701 & 0.669 & 0.731 & 0.788\\
    EasyControl~\cite{easycontrol} & 0.854 & 0.830 & 0.807 & 0.765 & 0.729 & 0.783 & 0.818\\
    UNO~\cite{UNO} & 0.737 & 0.763 & 0.730 & 0.674 & 0.628 & 0.699 & 0.718\\
    FLUX.1-Kontext-dev~\cite{flux1} & 0.849 & 0.813 & 0.778 & 0.731 & 0.694 & 0.754 & 0.802\\
    FLUX.2-dev~\cite{flux2} & 0.928 & 0.886 & 0.863 & 0.823 & 0.793 & 0.841 & 0.885\\
    NanobananaPro~\cite{nano} & \textbf{0.936} & 0.894 & 0.865 & 0.827 & 0.807 & 0.848 & 0.892\\
    Qwen-Editor~\cite{qweneditor} & \textbf{0.936} & \underline{0.903} & \underline{0.876} & \underline{0.840} & \underline{0.819} & \underline{0.859} & \underline{0.898}\\
    HuiWa~\cite{huiwa} & \underline{0.933} & 0.882 & 0.859 & 0.821 & 0.793 & 0.839 & 0.886\\
    YingHui~\cite{yinghui} & \textbf{0.936} & \textbf{0.904} & \textbf{0.882} & \textbf{0.847} & \textbf{0.823} & \textbf{0.864} & \textbf{0.900}\\
    \bottomrule
    \end{tabular}
    }
    \vspace{-10pt}
\end{wraptable}%

\textbf{Robustness to Erosion.}~
A decreasing trend in similarity scores is observed for all methods as the mask erodes. However, the rate of decay varies. OOTD drops significantly from $0.797$ to $0.669$ ($\Delta \approx 0.13$), implying its high scores rely partly on boundary correctness. 
It is also worth noting the foundational leap between generations: FLUX.2-dev significantly outperforms FLUX.1-Kontext-dev across all levels ($\bar{\mathcal{S}}_{rep}: 0.841$ vs. $0.754$), underscoring the critical role of a strong generative backbone in preserving high-frequency details.
In contrast, leading commercial solutions like YingHui maintain high fidelity even at the deepest erosion level ($\mathcal{S}^{(3)}_{rep}=0.823$), demonstrating the successful learning of valid internal texture representations rather than merely relying on edge completion.
\setlength{\columnsep}{15pt} 
\begin{wrapfigure}{r}{0.48\textwidth} 
    \vspace{-15pt} 
    \centering
    \includegraphics[width=\linewidth]{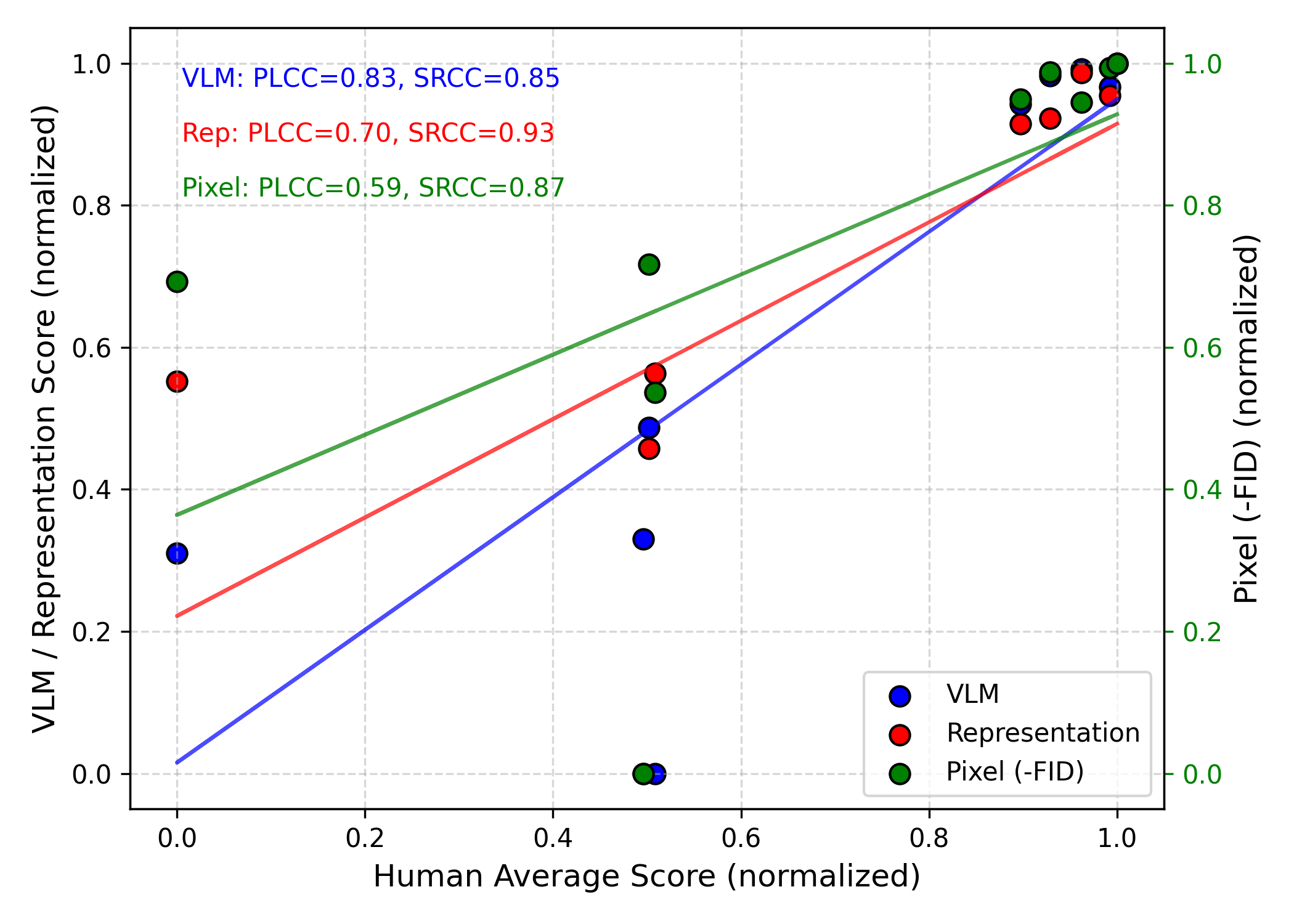}
    
    \vspace{-10pt}
    \small 
    \caption{\textbf{Correlation Analysis.} Scatter plot comparing objective metrics with human ratings. Both our \textbf{Representation} (Red) and \textbf{VLM} (Blue) metrics show strong positive correlations.}
    \label{fig:correlation}
    \vspace{-20pt} 
\end{wrapfigure}

\textbf{Global \& Local Consistency.}~
While NanobananaPro and Qwen-Editor achieve competitive scores in global consistency ($\mathcal{S}_{global} = 0.936$), YingHui consistently outperforms them in local garment similarity ($\bar{\mathcal{S}}_{rep} = 0.864$). This highlights the limitation of using whole-image embeddings for try-on assessment: a model can generate a visually pleasing image (high global score) while failing to preserve the specific details of the merchandise (lower local score).
Consequently, judging by the aggregated metric $\bar{\mathcal{S}}$, modern deployment-oriented systems have effectively bridged the gap between semantic coherence and pixel-level fidelity, establishing a clear upper bound that leaves earlier open-source attempts (e.g., UNO, OOTD) distinctly behind.

\subsection{Pixel-based Metrics and Correlation Analysis}
\label{sec:correlation}
Finally, we report standard pixel-level metrics (PSNR, SSIM, LPIPS, FID) in Table~\ref{tab:pixel_eval} and conduct a meta-evaluation of all metrics against human judgment in Table~\ref{tab:correlation_core}.
\paragraph{Saturation of Traditional Metrics.}
In Table~\ref{tab:pixel_eval}, \textbf{Qwen-Editor} achieves the best scores in PSNR ($26.343$) and SSIM ($0.905$). However, visual inspection suggests that it often minimizes pixel-wise error by smoothing textures. Conversely, \textbf{YingHui} achieves the best FID ($7.372$), indicating superior distribution-level realism.
The discrepancy between high PSNR and lower perceptual quality underscores the limitations of pixel-wise metrics for generative tasks.
\paragraph{Meta-Evaluation: Ranking Consistency.}
We compare model-level metric scores with human scores under the specified rubric using Spearman ($\rho_s$), Pearson ($\rho_p$), and Kendall's Tau ($\rho_k$).
As shown in Table~\ref{tab:correlation_core} and Figure~\ref{fig:correlation}, our \textbf{Representation Metric ($\bar{\mathcal{S}}$)} achieves the highest rank correlations among the evaluated metrics, with $\rho_k=0.833$ and $\rho_s=0.933$.
The higher Kendall's Tau than SSIM ($\rho_k=0.611$) indicates closer agreement with human model rankings in this evaluation.

\paragraph{Disagreement and Color Consistency.}
Across 45 model-by-dimension mean-score comparisons (nine models, five dimensions), the VLM--human mean absolute error is 0.299 on the 1--5 scale, with 40 comparisons within 0.5 points (Appendix~\ref{app:vlm_human_disagreement}).
UNO accounts for four of the five larger discrepancies, with overestimation of background, identity/body, texture, and shape preservation.
A separate CIEDE2000 analysis of generated test outputs measures garment color differences using both region-mean and pixel-wise $\Delta E_{00}$ (Appendix~\ref{app:color_consistency}).
YingHui achieves the lowest mean pixel-wise garment color error, despite the local color shift visible in Figure~\ref{fig:qualitative_more}.
These findings support model-ranking agreement under the shared rubric while motivating separate color diagnostics for errors that aggregate scores can obscure.

Furthermore, extended qualitative comparisons across complex scenarios and additional evaluations on supplementary baselines (e.g., DCI-VTON, PromptDresser) are provided in Figure~\ref{fig:qualitative_more} and Appendix~\ref{app:more_baselines}, respectively.
\begin{table}[htbp]
    \centering
    \begin{minipage}[t]{0.48\textwidth}
        \centering
        \caption{\textbf{Pixel-based Evaluation.}
        While Qwen-Editor dominates in pixel-alignment metrics (PSNR/SSIM), YingHui achieves the best FID, indicating superior distribution-level realism. $\uparrow$ denotes higher is better; $\downarrow$ denotes lower is better.}
        \label{tab:pixel_eval}
        
        \vspace{4pt} 
        \resizebox{\linewidth}{!}{
        \begin{tabular}{lcccc}
        \toprule
        \textbf{Method} & \textbf{PSNR} $\uparrow$ & \textbf{SSIM} $\uparrow$ & \textbf{LPIPS} $\downarrow$ & \textbf{FID} $\downarrow$ \\
        \midrule
        OOTD~\cite{ootdiffusion} & 16.534 & 0.794 & 0.255 & 55.216 \\
        EasyControl~\cite{easycontrol} & 15.731 & 0.779 & 0.277 & 39.079 \\
        UNO~\cite{UNO} & 12.096 & 0.726 & 0.417 & 110.534 \\
        FLUX.1-Kontext-dev~\cite{flux1} & 15.389 & 0.747 & 0.280 & 36.555 \\
        FLUX.2-dev~\cite{flux2} & 22.884 & 0.873 & 0.117 & 12.562 \\
        NanobananaPro~\cite{nano} & \underline{24.681} & \underline{0.890} & \underline{0.089} & \underline{7.989} \\
        Qwen-Editor~\cite{qweneditor} & \textbf{26.343} & \textbf{0.905} & \textbf{0.082} & 13.037 \\
        HuiWa~\cite{huiwa} & 23.045 & 0.875 & 0.100 & 8.619 \\
        YingHui~\cite{yinghui} & 22.593 & 0.870 & 0.101 & \textbf{7.372} \\
        \bottomrule
        \end{tabular}
        }
    \end{minipage}
    \hfill 
    \begin{minipage}[t]{0.48\textwidth}
        \centering
        \caption{\textbf{Meta-Evaluation: Correlation with Human Judgment.}
        Correlations are computed across model-level scores under the shared rubric. $\bar{\mathcal{S}}$ achieves the highest Kendall's Tau ($\rho_k$) among the evaluated metrics.}
        \label{tab:correlation_core}
        
        \vspace{4pt}
        \small 
        \begin{tabular}{lccc}
        \toprule
        \textbf{Metric} & $\rho_s \uparrow$ & $\rho_k \uparrow$ & $\rho_p \uparrow$ \\
        \midrule
        \textbf{$s_{avg}$ (VLM)} & 0.850 & \underline{0.722} & \textbf{0.828} \\
        \textbf{$\bar{\mathcal{S}}$ (Rep)} & \textbf{0.933} & \textbf{0.833} & 0.701 \\
        \midrule
        PSNR & 0.767 & 0.611 & \underline{0.819} \\
        SSIM & 0.767 & 0.611 & 0.801 \\
        $-$LPIPS & 0.833 & 0.667 & 0.782 \\
        $-$FID & \underline{0.867} & \underline{0.722} & 0.588 \\
        \bottomrule
        \end{tabular}
    \end{minipage}
\end{table}
\section{Conclusion}
In this paper, we introduce \textbf{OpenVTON-Bench}, an application-level benchmark designed to bridge the gap between generative capability and rigorous evaluation in Virtual Try-On. By leveraging DINOv3-based semantic clustering and Gemini-powered dense captioning, we construct 100K high-resolution ($1.5$K) pairs that effectively mitigate the ``studio-centric bias'' of prior works. We further establish a hybrid evaluation protocol combining VLM semantic reasoning with a structure-aware Multi-Scale Representation Metric. Our benchmarking reveals a notable ``texture-realism gap'' in state-of-the-art diffusion models: while photorealistic, they often hallucinate fine-grained garment details—a nuance our metric disentangles more effectively than traditional pixel-level measures.

\textbf{Limitations. }Our pipeline relies on off-the-shelf foundation models (e.g., Gemini, DINOv3) for data filtering and captioning, which may introduce semantic biases or hallucinations. Despite the increased resolution, complex multi-layer occlusions and extreme poses remain under-represented. The protocol assesses 2D visual fidelity and plausibility; it does not validate physical fit, garment sizing or grading, or material-draping physics. Future work will refine automated annotations and expand pose and garment diversity. The public code and dataset support further research on VTON systems that preserve garment details while producing plausible images.

\clearpage
\begin{ack}
This work received no external funding. The authors declare no competing interests.
\end{ack}
%
\nocite{gou2023taming, Qwen-VL, grounddino}
\bibliography{paper}

\begin{thebibliography}{63}
\providecommand{\natexlab}[1]{#1}
\providecommand{\url}[1]{\texttt{#1}}
\expandafter\ifx\csname urlstyle\endcsname\relax
  \providecommand{\doi}[1]{doi: #1}\else
  \providecommand{\doi}{doi: \begingroup \urlstyle{rm}\Url}\fi

\bibitem[Bai et~al.(2023)Bai, Bai, Yang, Wang, Tan, Wang, Lin, Zhou, and
  Zhou]{Qwen-VL}
Jinze Bai, Shuai Bai, Shusheng Yang, Shijie Wang, Sinan Tan, Peng Wang, Junyang
  Lin, Chang Zhou, and Jingren Zhou.
\newblock Qwen-vl: A versatile vision-language model for understanding,
  localization, text reading, and beyond.
\newblock \emph{arXiv preprint arXiv:2308.12966}, 2023.

\bibitem[Bai et~al.(2025)Bai, Chen, Liu, Wang, Ge, Song, Dang, Wang, Wang,
  Tang, et~al.]{bai2025qwen2}
Shuai Bai, Keqin Chen, Xuejing Liu, Jialin Wang, Wenbin Ge, Sibo Song, Kai
  Dang, Peng Wang, Shijie Wang, Jun Tang, et~al.
\newblock Qwen2. 5-vl technical report.
\newblock \emph{arXiv preprint arXiv:2502.13923}, 2025.

\bibitem[Carion et~al.(2025)Carion, Gustafson, Hu, Debnath, Hu, Suris, Ryali,
  Alwala, Khedr, Huang, Lei, Ma, Guo, Kalla, Marks, Greer, Wang, Sun, Rädle,
  Afouras, Mavroudi, Xu, Wu, Zhou, Momeni, Hazra, Ding, Vaze, Porcher, Li, Li,
  Kamath, Cheng, Dollár, Ravi, Saenko, Zhang, and Feichtenhofer]{sam3}
Nicolas Carion, Laura Gustafson, Yuan-Ting Hu, Shoubhik Debnath, Ronghang Hu,
  Didac Suris, Chaitanya Ryali, Kalyan~Vasudev Alwala, Haitham Khedr, Andrew
  Huang, Jie Lei, Tengyu Ma, Baishan Guo, Arpit Kalla, Markus Marks, Joseph
  Greer, Meng Wang, Peize Sun, Roman Rädle, Triantafyllos Afouras, Effrosyni
  Mavroudi, Katherine Xu, Tsung-Han Wu, Yu~Zhou, Liliane Momeni, Rishi Hazra,
  Shuangrui Ding, Sagar Vaze, Francois Porcher, Feng Li, Siyuan Li, Aishwarya
  Kamath, Ho~Kei Cheng, Piotr Dollár, Nikhila Ravi, Kate Saenko, Pengchuan
  Zhang, and Christoph Feichtenhofer.
\newblock Sam 3: Segment anything with concepts, 2025.
\newblock URL \url{https://arxiv.org/abs/2511.16719}.

\bibitem[Chen et~al.(2024)Chen, Chen, Zhang, Wang, Liu, Zhou, Zhang, Wan, Zhou,
  and Sun]{chen2024mllm}
Dongping Chen, Ruoxi Chen, Shilin Zhang, Yaochen Wang, Yinuo Liu, Huichi Zhou,
  Qihui Zhang, Yao Wan, Pan Zhou, and Lichao Sun.
\newblock Mllm-as-a-judge: Assessing multimodal llm-as-a-judge with
  vision-language benchmark.
\newblock In \emph{Forty-first International Conference on Machine Learning},
  2024.

\bibitem[Choi et~al.(2021)Choi, Park, Lee, and Choo]{choi2021viton}
Seunghwan Choi, Sunghyun Park, Minsoo Lee, and Jaegul Choo.
\newblock Viton-hd: High-resolution virtual try-on via misalignment-aware
  normalization.
\newblock In \emph{Proceedings of the IEEE/CVF conference on computer vision
  and pattern recognition}, pages 14131--14140, 2021.

\bibitem[Chong et~al.(2024)Chong, Dong, Li, Zhang, Zhang, Zhang, Zhao, Jiang,
  and Liang]{catvton2024}
Zheng Chong, Xiao Dong, Haoxiang Li, Shiyue Zhang, Wenqing Zhang, Xujie Zhang,
  Hanqing Zhao, Dongmei Jiang, and Xiaodan Liang.
\newblock Catvton: Concatenation is all you need for virtual try-on with
  diffusion models.
\newblock \emph{arXiv preprint arXiv:2407.15886}, 2024.

\bibitem[Chong et~al.(2025)Chong, Zhang, Zhang, Zheng, Dong, Li, Wu, Jiang, and
  Liang]{catv2ton2025}
Zheng Chong, Wenqing Zhang, Shiyue Zhang, Jun Zheng, Xiao Dong, Haoxiang Li,
  Yiling Wu, Dongmei Jiang, and Xiaodan Liang.
\newblock Catv2ton: Taming diffusion transformers for vision-based virtual
  try-on with temporal concatenation.
\newblock \emph{arXiv preprint arXiv:2501.11325}, 2025.

\bibitem[Cui et~al.(2024)Cui, Mahajan, Shah, Gomathinayagam, Liu, and
  Lazebnik]{StreetTryon}
Aiyu Cui, Jay Mahajan, Viraj Shah, Preeti Gomathinayagam, Chang Liu, and
  Svetlana Lazebnik.
\newblock Street tryon: Learning in-the-wild virtual try-on from unpaired
  person images.
\newblock In \emph{Proceedings of the IEEE/CVF Conference on Computer Vision
  and Pattern Recognition}, pages 8235--8239, 2024.

\bibitem[Dong et~al.(2019)Dong, Liang, Wang, Lai, Zhu, and Yin]{MPV}
Haoye Dong, Xiaodan Liang, Bochao Wang, Hanjiang Lai, Jia Zhu, and Jian Yin.
\newblock Towards multi-pose guided virtual try-on network, 2019.
\newblock URL \url{https://arxiv.org/abs/1902.11026}.

\bibitem[Duchon(2006)]{TPS}
Jean Duchon.
\newblock Splines minimizing rotation-invariant semi-norms in sobolev spaces.
\newblock In \emph{Constructive theory of functions of several variables:
  proceedings of a conference held at oberwolfach April 25--May 1, 1976}, pages
  85--100. Springer, 2006.

\bibitem[Feng et~al.(2022)Feng, Ma, Shen, Gao, Liu, Li, Ou, Zhao, and Zha]{ESF}
Ruili Feng, Cheng Ma, Chengji Shen, Xin Gao, Zhenjiang Liu, Xiaobo Li, Kairi
  Ou, Deli Zhao, and Zheng-Jun Zha.
\newblock Weakly supervised high-fidelity clothing model generation.
\newblock In \emph{Proceedings of the IEEE/CVF conference on computer vision
  and pattern recognition}, pages 3440--3449, 2022.

\bibitem[Fu et~al.(2022)Fu, Li, Jiang, Lin, Qian, Loy, Wu, and Liu]{SHHQ}
Jianglin Fu, Shikai Li, Yuming Jiang, Kwan-Yee Lin, Chen Qian, Chen~Change Loy,
  Wayne Wu, and Ziwei Liu.
\newblock Stylegan-human: A data-centric odyssey of human generation, 2022.
\newblock URL \url{https://arxiv.org/abs/2204.11823}.

\bibitem[Ge et~al.(2021)Ge, Song, Ge, Yang, Liu, and Luo]{ge2021disentangled}
Chongjian Ge, Yibing Song, Yuying Ge, Han Yang, Wei Liu, and Ping Luo.
\newblock Disentangled cycle consistency for highly-realistic virtual try-on.
\newblock In \emph{Proceedings of the IEEE/CVF conference on computer vision
  and pattern recognition}, pages 16928--16937, 2021.

\bibitem[{Google DeepMind}(2025)]{nano}
{Google DeepMind}.
\newblock Gemini 3 pro image model card.
\newblock Technical report, Google, 2025.
\newblock URL
  \url{https://storage.googleapis.com/deepmind-media/Model-Cards/Gemini-3-Pro-Image-Model-Card.pdf}.

\bibitem[Gou et~al.(2023)Gou, Sun, Zhang, Si, Qian, and Zhang]{gou2023taming}
Junhong Gou, Siyu Sun, Jianfu Zhang, Jianlou Si, Chen Qian, and Liqing Zhang.
\newblock Taming the power of diffusion models for high-quality virtual try-on
  with appearance flow.
\newblock In \emph{Proceedings of the 31st ACM international conference on
  multimedia}, pages 7599--7607, 2023.

\bibitem[Han et~al.(2018)Han, Wu, Wu, Yu, and Davis]{VITON}
Xintong Han, Zuxuan Wu, Zhe Wu, Ruichi Yu, and Larry~S. Davis.
\newblock Viton: An image-based virtual try-on network, 2018.
\newblock URL \url{https://arxiv.org/abs/1711.08447}.

\bibitem[Heusel et~al.(2017)Heusel, Ramsauer, Unterthiner, Nessler, and
  Hochreiter]{FID}
Martin Heusel, Hubert Ramsauer, Thomas Unterthiner, Bernhard Nessler, and Sepp
  Hochreiter.
\newblock Gans trained by a two time-scale update rule converge to a local nash
  equilibrium.
\newblock \emph{Advances in neural information processing systems}, 30, 2017.

\bibitem[Honda(2019)]{VITON-GAN}
Shion Honda.
\newblock Viton-gan: Virtual try-on image generator trained with adversarial
  loss.
\newblock \emph{Eurographics 2019 - Posters}, 2019.
\newblock \doi{10.2312/EGP.20191043}.
\newblock URL \url{https://diglib.eg.org/handle/10.2312/egp20191043}.

\bibitem[Hore and Ziou(2010)]{PSNR}
Alain Hore and Djemel Ziou.
\newblock Image quality metrics: Psnr vs. ssim.
\newblock In \emph{2010 20th international conference on pattern recognition},
  pages 2366--2369. IEEE, 2010.

\bibitem[{HuiWa AI Creative and Marketing Platform}(2024)]{huiwa}
{HuiWa AI Creative and Marketing Platform}.
\newblock Huiwa ai creative and marketing platform, 2024.
\newblock URL \url{https://www.ihuiwa.com/}.

\bibitem[Issenhuth et~al.(2020)Issenhuth, Mary, and
  Calauzenes]{issenhuth2020not}
Thibaut Issenhuth, J{\'e}r{\'e}mie Mary, and Cl{\'e}ment Calauzenes.
\newblock Do not mask what you do not need to mask: a parser-free virtual
  try-on.
\newblock In \emph{European Conference on Computer Vision}, pages 619--635.
  Springer, 2020.

\bibitem[Jaderberg et~al.(2015)Jaderberg, Simonyan, Zisserman, et~al.]{STN}
Max Jaderberg, Karen Simonyan, Andrew Zisserman, et~al.
\newblock Spatial transformer networks.
\newblock \emph{Advances in neural information processing systems}, 28, 2015.

\bibitem[Kim et~al.(2024)Kim, Gu, Park, Park, and Choo]{stableviton2024}
Jeongho Kim, Guojung Gu, Minho Park, Sunghyun Park, and Jaegul Choo.
\newblock Stableviton: Learning semantic correspondence with latent diffusion
  model for virtual try-on.
\newblock In \emph{Proceedings of the IEEE/CVF conference on computer vision
  and pattern recognition}, pages 8176--8185, 2024.

\bibitem[Kim et~al.(2025)Kim, Jin, Park, and Choo]{PromptDresser2025}
Jeongho Kim, Hoiyeong Jin, Sunghyun Park, and Jaegul Choo.
\newblock Promptdresser: Improving the quality and controllability of virtual
  try-on via generative textual prompt and prompt-aware mask, 2025.
\newblock URL \url{https://arxiv.org/abs/2412.16978}.

\bibitem[Ku et~al.(2024)Ku, Jiang, Wei, Yue, and Chen]{ku2024viescore}
Max Ku, Dongfu Jiang, Cong Wei, Xiang Yue, and Wenhu Chen.
\newblock {VIEScore}: Towards explainable metrics for conditional image
  synthesis evaluation.
\newblock In \emph{Proceedings of the 62nd Annual Meeting of the Association
  for Computational Linguistics (Volume 1: Long Papers)}, pages 12268--12290.
  Association for Computational Linguistics, 2024.
\newblock \doi{10.18653/v1/2024.acl-long.663}.
\newblock URL \url{https://aclanthology.org/2024.acl-long.663/}.

\bibitem[Labs(2025)]{flux2}
Black~Forest Labs.
\newblock {FLUX.2: Frontier Visual Intelligence}.
\newblock \url{https://bfl.ai/blog/flux-2}, 2025.

\bibitem[Labs et~al.(2025)Labs, Batifol, Blattmann, Boesel, Consul, Diagne,
  Dockhorn, English, English, Esser, Kulal, Lacey, Levi, Li, Lorenz, Müller,
  Podell, Rombach, Saini, Sauer, and Smith]{flux1}
Black~Forest Labs, Stephen Batifol, Andreas Blattmann, Frederic Boesel, Saksham
  Consul, Cyril Diagne, Tim Dockhorn, Jack English, Zion English, Patrick
  Esser, Sumith Kulal, Kyle Lacey, Yam Levi, Cheng Li, Dominik Lorenz, Jonas
  Müller, Dustin Podell, Robin Rombach, Harry Saini, Axel Sauer, and Luke
  Smith.
\newblock Flux.1 kontext: Flow matching for in-context image generation and
  editing in latent space, 2025.
\newblock URL \url{https://arxiv.org/abs/2506.15742}.

\bibitem[Lee et~al.(2022)Lee, Gu, Park, Choi, and Choo]{HR-VITON}
Sangyun Lee, Gyojung Gu, Sunghyun Park, Seunghwan Choi, and Jaegul Choo.
\newblock High-resolution virtual try-on with misalignment and
  occlusion-handled conditions.
\newblock In \emph{European Conference on Computer Vision}, pages 204--219.
  Springer, 2022.

\bibitem[Lewis et~al.(2021)Lewis, Varadharajan, and
  Kemelmacher-Shlizerman]{TryOnGAN}
Kathleen~M Lewis, Srivatsan Varadharajan, and Ira Kemelmacher-Shlizerman.
\newblock Tryongan: Body-aware try-on via layered interpolation, 2021.
\newblock URL \url{https://arxiv.org/abs/2101.02285}.

\bibitem[Li et~al.(2024)Li, Liu, Singh, Wang, Zhang, Plummer, and Lin]{LH-400K}
Nannan Li, Qing Liu, Krishna~Kumar Singh, Yilin Wang, Jianming Zhang, Bryan~A
  Plummer, and Zhe Lin.
\newblock Unihuman: A unified model for editing human images in the wild.
\newblock In \emph{Proceedings of the IEEE/CVF conference on computer vision
  and pattern recognition}, pages 2039--2048, 2024.

\bibitem[Li et~al.(2019)Li, Huang, and Loy]{flownet}
Yining Li, Chen Huang, and Chen~Change Loy.
\newblock Dense intrinsic appearance flow for human pose transfer.
\newblock In \emph{Proceedings of the IEEE/CVF conference on computer vision
  and pattern recognition}, pages 3693--3702, 2019.

\bibitem[Lin et~al.(2025)Lin, Hu, Li, Geng, Koh, Zettlemoyer, Althoff, and
  Ghazvininejad]{lin2025self}
Inna~Wanyin Lin, Yushi Hu, Shuyue~Stella Li, Scott Geng, Pang~Wei Koh, Luke
  Zettlemoyer, Tim Althoff, and Marjan Ghazvininejad.
\newblock Self-improving vlm judges without human annotations.
\newblock \emph{arXiv preprint arXiv:2512.05145}, 2025.

\bibitem[Liu et~al.(2023)Liu, Zeng, Ren, Li, Zhang, Yang, Li, Yang, Su, Zhu,
  et~al.]{grounddino}
Shilong Liu, Zhaoyang Zeng, Tianhe Ren, Feng Li, Hao Zhang, Jie Yang, Chunyuan
  Li, Jianwei Yang, Hang Su, Jun Zhu, et~al.
\newblock Grounding dino: Marrying dino with grounded pre-training for open-set
  object detection.
\newblock \emph{arXiv preprint arXiv:2303.05499}, 2023.

\bibitem[Liu et~al.(2016)Liu, Luo, Qiu, Wang, and Tang]{deepfashion}
Ziwei Liu, Ping Luo, Shi Qiu, Xiaogang Wang, and Xiaoou Tang.
\newblock Deepfashion: Powering robust clothes recognition and retrieval with
  rich annotations.
\newblock In \emph{Proceedings of IEEE Conference on Computer Vision and
  Pattern Recognition (CVPR)}, June 2016.

\bibitem[Luo et~al.(2001)Luo, Cui, and Rigg]{luo2001ciede2000}
M.~R. Luo, G.~Cui, and B.~Rigg.
\newblock The development of the {CIE} 2000 colour-difference formula:
  {CIEDE2000}.
\newblock \emph{Color Research \& Application}, 26\penalty0 (5):\penalty0
  340--350, 2001.
\newblock \doi{10.1002/col.1049}.

\bibitem[Morelli et~al.(2022)Morelli, Fincato, Cornia, Landi, Cesari, and
  Cucchiara]{DressCode}
Davide Morelli, Matteo Fincato, Marcella Cornia, Federico Landi, Fabio Cesari,
  and Rita Cucchiara.
\newblock Dress code: High-resolution multi-category virtual try-on, 2022.
\newblock URL \url{https://arxiv.org/abs/2204.08532}.

\bibitem[Morelli et~al.(2023)Morelli, Baldrati, Cartella, Cornia, Bertini, and
  Cucchiara]{LaDIVTON2023}
Davide Morelli, Alberto Baldrati, Giuseppe Cartella, Marcella Cornia, Marco
  Bertini, and Rita Cucchiara.
\newblock Ladi-vton: Latent diffusion textual-inversion enhanced virtual
  try-on.
\newblock In \emph{Proceedings of the 31st ACM international conference on
  multimedia}, pages 8580--8589, 2023.

\bibitem[Schuhmann et~al.(2021)Schuhmann, Vencu, Beaumont, Kaczmarczyk, Mullis,
  Katta, Coombes, Jitsev, and Komatsuzaki]{LAION}
Christoph Schuhmann, Richard Vencu, Romain Beaumont, Robert Kaczmarczyk,
  Clayton Mullis, Aarush Katta, Theo Coombes, Jenia Jitsev, and Aran
  Komatsuzaki.
\newblock Laion-400m: Open dataset of clip-filtered 400 million image-text
  pairs, 2021.
\newblock URL \url{https://arxiv.org/abs/2111.02114}.

\bibitem[Shen et~al.(2024)Shen, Jiang, He, Ye, Wang, Du, Li, and
  Tang]{IMAGDressing-v1}
Fei Shen, Xin Jiang, Xin He, Hu~Ye, Cong Wang, Xiaoyu Du, Zechao Li, and Jinhui
  Tang.
\newblock Imagdressing-v1: Customizable virtual dressing, 2024.
\newblock URL \url{https://arxiv.org/abs/2407.12705}.

\bibitem[Shim et~al.(2024)Shim, Chung, and Heo]{SDVTON2024}
Sang-Heon Shim, Jiwoo Chung, and Jae-Pil Heo.
\newblock Towards squeezing-averse virtual try-on via sequential deformation.
\newblock In \emph{Proceedings of the AAAI Conference on Artificial
  Intelligence}, volume~38, pages 4856--4863, 2024.

\bibitem[Sim{\'e}oni et~al.(2025)Sim{\'e}oni, Vo, Seitzer, Baldassarre, Oquab,
  Jose, Khalidov, Szafraniec, Yi, Ramamonjisoa, Massa, Haziza, Wehrstedt, Wang,
  Darcet, Moutakanni, Sentana, Roberts, Vedaldi, Tolan, Brandt, Couprie,
  Mairal, J{\'e}gou, Labatut, and Bojanowski]{dinov3}
Oriane Sim{\'e}oni, Huy~V. Vo, Maximilian Seitzer, Federico Baldassarre, Maxime
  Oquab, Cijo Jose, Vasil Khalidov, Marc Szafraniec, Seungeun Yi, Micha{\"e}l
  Ramamonjisoa, Francisco Massa, Daniel Haziza, Luca Wehrstedt, Jianyuan Wang,
  Timoth{\'e}e Darcet, Th{\'e}o Moutakanni, Leonel Sentana, Claire Roberts,
  Andrea Vedaldi, Jamie Tolan, John Brandt, Camille Couprie, Julien Mairal,
  Herv{\'e} J{\'e}gou, Patrick Labatut, and Piotr Bojanowski.
\newblock {DINOv3}, 2025.
\newblock URL \url{https://arxiv.org/abs/2508.10104}.

\bibitem[Song et~al.(2024)Song, Zhang, Zhou, Nie, Tong, Kankanhalli, and
  Liu]{song2024imagebasedvirtualtryonsurvey}
Dan Song, Xuanpu Zhang, Juan Zhou, Weizhi Nie, Ruofeng Tong, Mohan Kankanhalli,
  and An-An Liu.
\newblock Image-based virtual try-on: A survey, 2024.
\newblock URL \url{https://arxiv.org/abs/2311.04811}.

\bibitem[Sun et~al.(2025)Sun, Hong, Zhan, Lan, Zhu, Wang, Zhang, and
  Zhang]{dsvton2025}
Xianbing Sun, Yan Hong, Jiahui Zhan, Jun Lan, Huijia Zhu, Weiqiang Wang, Liqing
  Zhang, and Jianfu Zhang.
\newblock Ds-vton: High-quality virtual try-on via disentangled dual-scale
  generation.
\newblock \emph{arXiv preprint arXiv:2506.00908}, 2025.

\bibitem[Team(2025)]{gemini}
Gemini Team.
\newblock Gemini: A family of highly capable multimodal models, 2025.
\newblock URL \url{https://arxiv.org/abs/2312.11805}.

\bibitem[Wan et~al.(2025)Wan, Chen, Cai, Pan, Yao, and Mei]{wan2025vtonvllm}
Siqi Wan, Jingwen Chen, Qi~Cai, Yingwei Pan, Ting Yao, and Tao Mei.
\newblock {VTON-VLLM}: Aligning virtual try-on models with human preferences.
\newblock In \emph{Advances in Neural Information Processing Systems},
  volume~38, pages 147641--147664. Curran Associates, Inc., 2025.
\newblock \doi{10.52202/085713-4937}.
\newblock URL
  \url{https://proceedings.neurips.cc/paper_files/paper/2025/hash/d8bb4be0e26f0d819043d33a945e8c0d-Abstract-Conference.html}.

\bibitem[Wang et~al.(2018)Wang, Zheng, Liang, Chen, Lin, and Yang]{CP-VTON}
Bochao Wang, Huabin Zheng, Xiaodan Liang, Yimin Chen, Liang Lin, and Meng Yang.
\newblock Toward characteristic-preserving image-based virtual try-on network,
  2018.
\newblock URL \url{https://arxiv.org/abs/1807.07688}.

\bibitem[Wang et~al.(2025)Wang, Zhang, Di, Zhang, and Zuo]{wang2025mv}
Haoyu Wang, Zhilu Zhang, Donglin Di, Shiliang Zhang, and Wangmeng Zuo.
\newblock Mv-vton: Multi-view virtual try-on with diffusion models.
\newblock In \emph{Proceedings of the AAAI Conference on Artificial
  Intelligence}, volume~39, pages 7682--7690, 2025.

\bibitem[Wang et~al.(2004)Wang, Bovik, Sheikh, and Simoncelli]{SSIM}
Zhou Wang, Alan~C. Bovik, Hamid~R. Sheikh, and Eero~P. Simoncelli.
\newblock Image quality assessment: From error visibility to structural
  similarity.
\newblock \emph{IEEE Transactions on Image Processing}, 13\penalty0
  (4):\penalty0 600--612, 2004.
\newblock \doi{10.1109/TIP.2003.819861}.

\bibitem[Wei et~al.(2026)Wei, Wu, Xu, Li, Duan, Min, and Zhai]{wei2024vtonqa}
Xinyi Wei, Sijing Wu, Zitong Xu, Yunhao Li, Huiyu Duan, Xiongkuo Min, and
  Guangtao Zhai.
\newblock Vtonqa: A multi-dimensional quality assessment dataset for virtual
  try-on, 2026.
\newblock URL \url{https://arxiv.org/abs/2601.02945}.

\bibitem[Wu et~al.(2025{\natexlab{a}})Wu, Li, Zhou, Lin, Gao, Yan, ming Yin,
  Bai, Xu, Chen, Chen, Tang, Zhang, Wang, Yang, Yu, Cheng, Liu, Li, Zhang,
  Meng, Wei, Ni, Chen, Cao, Peng, Qu, Wu, Wang, Yu, Wen, Feng, Xu, Wang, Zhang,
  Zhu, Wu, Cai, and Liu]{qweneditor}
Chenfei Wu, Jiahao Li, Jingren Zhou, Junyang Lin, Kaiyuan Gao, Kun Yan, Sheng
  ming Yin, Shuai Bai, Xiao Xu, Yilei Chen, Yuxiang Chen, Zecheng Tang, Zekai
  Zhang, Zhengyi Wang, An~Yang, Bowen Yu, Chen Cheng, Dayiheng Liu, Deqing Li,
  Hang Zhang, Hao Meng, Hu~Wei, Jingyuan Ni, Kai Chen, Kuan Cao, Liang Peng,
  Lin Qu, Minggang Wu, Peng Wang, Shuting Yu, Tingkun Wen, Wensen Feng,
  Xiaoxiao Xu, Yi~Wang, Yichang Zhang, Yongqiang Zhu, Yujia Wu, Yuxuan Cai, and
  Zenan Liu.
\newblock Qwen-image technical report, 2025{\natexlab{a}}.
\newblock URL \url{https://arxiv.org/abs/2508.02324}.

\bibitem[Wu et~al.(2025{\natexlab{b}})Wu, Huang, Wu, Cheng, Ding, and He]{UNO}
Shaojin Wu, Mengqi Huang, Wenxu Wu, Yufeng Cheng, Fei Ding, and Qian He.
\newblock Less-to-more generalization: Unlocking more controllability by
  in-context generation.
\newblock \emph{arXiv preprint arXiv:2504.02160}, 2025{\natexlab{b}}.

\bibitem[Xiaobin et~al.(2025)Xiaobin, Yujie, Donghao, Xu, Jiangning, Junwei,
  Chengjie, and Yanwei]{VTBench2025}
Hu~Xiaobin, Liang Yujie, Luo Donghao, Peng Xu, Zhang Jiangning, Zhu Junwei,
  Wang Chengjie, and Fu~Yanwei.
\newblock Vtbench: Comprehensive benchmark suite towards real-world virtual
  try-on models, 2025.
\newblock URL \url{https://arxiv.org/abs/2505.19571}.

\bibitem[Xie et~al.(2021)Xie, Huang, Zhao, Dong, Kampffmeyer, and Liang]{UPT}
Zhenyu Xie, Zaiyu Huang, Fuwei Zhao, Haoye Dong, Michael Kampffmeyer, and
  Xiaodan Liang.
\newblock Towards scalable unpaired virtual try-on via patch-routed
  spatially-adaptive gan, 2021.
\newblock URL \url{https://arxiv.org/abs/2111.10544}.

\bibitem[Xu et~al.(2025)Xu, Gu, Chen, and Chen]{ootdiffusion}
Yuhao Xu, Tao Gu, Weifeng Chen, and Arlene Chen.
\newblock Ootdiffusion: Outfitting fusion based latent diffusion for
  controllable virtual try-on.
\newblock In \emph{Proceedings of the AAAI Conference on Artificial
  Intelligence}, volume~39, pages 8996--9004, 2025.

\bibitem[Yang et~al.(2024)Yang, Ding, Hong, Huang, Tao, and
  Xu]{yang2024texture}
Xu~Yang, Changxing Ding, Zhibin Hong, Junhao Huang, Jin Tao, and Xiangmin Xu.
\newblock Texture-preserving diffusion models for high-fidelity virtual try-on.
\newblock In \emph{Proceedings of the IEEE/CVF conference on computer vision
  and pattern recognition}, pages 7017--7026, 2024.

\bibitem[Yang et~al.(2025)Yang, Li, He, Li, Xu, Dong, and Du]{yang2025omnivton}
Zhaotong Yang, Yuhui Li, Shengfeng He, Xinzhe Li, Yangyang Xu, Junyu Dong, and
  Yong Du.
\newblock Omnivton: Training-free universal virtual try-on.
\newblock In \emph{Proceedings of the IEEE/CVF International Conference on
  Computer Vision}, pages 16702--16711, 2025.

\bibitem[{YingHui AI Creative Platform}(2025)]{yinghui}
{YingHui AI Creative Platform}.
\newblock Yinghui ai creative platform, 2025.
\newblock URL \url{https://www.yinghuigen.com/}.

\bibitem[Zeng et~al.(2024)Zeng, Song, Nie, Tian, Wang, and Liu]{catdm2024}
Jianhao Zeng, Dan Song, Weizhi Nie, Hongshuo Tian, Tongtong Wang, and An-An
  Liu.
\newblock Cat-dm: Controllable accelerated virtual try-on with diffusion model.
\newblock In \emph{Proceedings of the IEEE/CVF conference on computer vision
  and pattern recognition}, pages 8372--8382, 2024.

\bibitem[Zhang et~al.(2018)Zhang, Isola, Efros, Shechtman, and Wang]{LPIPS}
Richard Zhang, Phillip Isola, Alexei~A Efros, Eli Shechtman, and Oliver Wang.
\newblock The unreasonable effectiveness of deep features as a perceptual
  metric.
\newblock In \emph{Proceedings of the IEEE conference on computer vision and
  pattern recognition}, pages 586--595, 2018.

\bibitem[Zhang et~al.(2025)Zhang, Yuan, Song, Wang, and Liu]{easycontrol}
Yuxuan Zhang, Yirui Yuan, Yiren Song, Haofan Wang, and Jiaming Liu.
\newblock Easycontrol: Adding efficient and flexible control for diffusion
  transformer.
\newblock In \emph{Proceedings of the IEEE/CVF International Conference on
  Computer Vision}, pages 19513--19524, 2025.

\bibitem[Zheng et~al.(2024)Zheng, Zhao, Xu, Dong, and Liang]{zheng2024viton}
Jun Zheng, Fuwei Zhao, Youjiang Xu, Xin Dong, and Xiaodan Liang.
\newblock Viton-dit: Learning in-the-wild video try-on from human dance videos
  via diffusion transformers.
\newblock \emph{arXiv preprint arXiv:2405.18326}, 2024.

\bibitem[Zhu et~al.(2025)Zhu, Wang, Chen, Liu, Ye, Gu, Tian, Duan, Su, Shao,
  et~al.]{zhu2025internvl3}
Jinguo Zhu, Weiyun Wang, Zhe Chen, Zhaoyang Liu, Shenglong Ye, Lixin Gu, Hao
  Tian, Yuchen Duan, Weijie Su, Jie Shao, et~al.
\newblock Internvl3: Exploring advanced training and test-time recipes for
  open-source multimodal models.
\newblock \emph{arXiv preprint arXiv:2504.10479}, 2025.

\bibitem[Zhu et~al.(2023)Zhu, Yang, Zhu, Reda, Chan, Saharia, Norouzi, and
  Kemelmacher-Shlizerman]{TryOnDiffusion2023}
Luyang Zhu, Dawei Yang, Tyler Zhu, Fitsum Reda, William Chan, Chitwan Saharia,
  Mohammad Norouzi, and Ira Kemelmacher-Shlizerman.
\newblock Tryondiffusion: A tale of two unets.
\newblock In \emph{Proceedings of the IEEE/CVF conference on computer vision
  and pattern recognition}, pages 4606--4615, 2023.

\end{thebibliography}
\bibliographystyle{plainnat}

\newpage
\appendix
\onecolumn

\section*{Overview of Supplementary Material}

This supplementary material follows the structure of the main paper: data provenance first, metric definitions second, validation of the evaluation protocol third, and experimental evidence last.
\begin{itemize}
    \item \textbf{Appendix~\ref{app:data_details}: Dataset Details: Construction and Visualization.} We describe the social-media and e-commerce data sources, sampling across visual clusters, automated mask generation, Gemini-2.0-Flash captioning, and dataset visualizations.
    \item \textbf{Appendix~\ref{app:metrics}: Evaluation Metrics \& Protocols.} We define the VLM-as-a-Judge protocol and the representation-based metrics, including the multi-scale garment fidelity score $\mathcal{S}_{\text{rep}}^{(k)}$.
    \item \textbf{Appendix~\ref{app:protocol_validation}: Validation of Evaluation Protocol.} We report correlations and score differences between automated metrics and model-level human scores under the shared rubric, inter-rater reliability based on Krippendorff's Alpha, and a separate color consistency diagnostic.
    \item \textbf{Appendix~\ref{app:results}: Additional Experimental Results.} We provide implementation details, supplementary baseline results, fine-tuning results on OpenVTON-Bench, and extended qualitative comparisons.
    \item \textbf{Appendix~\ref{app:ethics_compute}: Ethics, Licenses, Human Study, and Compute.} We discuss broader impacts, safeguards, external asset licenses, human evaluation protocol, IRB or equivalent review, and compute resources.
\end{itemize}
\section{Dataset Details: Construction and Visualization}
\label{app:data_details}
\label{app:data_collection}

OpenVTON-Bench is designed around high-resolution, high-fidelity, and semantically diverse virtual try-on pairs. The construction pipeline includes multi-source candidate collection, sampling across DINOv3 K-Means visual clusters, automated triplet construction, dense captioning, and distributional inspection.

\subsection{Comparison with Existing Datasets}
\label{app:dataset_comparison}
To highlight the necessity of our proposed benchmark, Table~\ref{tab:dataset_comparison} provides a quantitative comparison between OpenVTON-Bench and existing representative virtual try-on datasets. As shown, legacy datasets either suffer from low resolution (e.g., $256\times192$) or lack the scale required for modern diffusion models. While recent efforts attempt to reach higher resolutions, they are mostly closed-source or limited to small quantities. In contrast, OpenVTON-Bench offers the largest open-source collection of high-fidelity image pairs (reaching up to $1536^2$), successfully bridging the gap between academic research and application-level application requirements.
\begin{table}[htbp]
    \centering
    \caption{\textbf{Comparison of Virtual Try-On Datasets.}
    We compare OpenVTON-Bench with representative datasets in terms of scale, resolution, and accessibility. Unlike prior benchmarks that are either low-resolution or closed-source, our benchmark offers the largest open-source collection of high-fidelity images with $\min(H, W) \geq 1024$. \textit{Open-Source}: $\checkmark$ denotes publicly available, $\times$ denotes closed.}
    \label{tab:dataset_comparison}
    \vspace{0.5em}
    \begin{tabular}{lcccc}
        \toprule
        \textbf{Dataset} & \textbf{Year} & \textbf{Quantity} & \textbf{Resolution} & \textbf{Open-Source} \\
        \midrule
        VITON~\cite{VITON}                & 2018 & 16,253  & $256 \times 192$                     & $\checkmark$ \\
        VITON-HD~\cite{choi2021viton}     & 2021 & 13,679  & $1024 \times 768$                   & $\checkmark$ \\
        DressCode~\cite{DressCode}        & 2022 & 53,792  & $1024 \times 768$                   & $\checkmark$ \\
        SHHQ~\cite{SHHQ}                  & 2022 & 231,176 & $1024 \times 768$                   & $\checkmark$ \\
        StreetTryOn~\cite{StreetTryon}    & 2024 & 14,453  & $512 \times 320$                    & $\checkmark$ \\
        LH-400K~\cite{LH-400K}            & 2024 & 409,270 & $512 \times 512$                    & $\checkmark$ \\
        VTBench~\cite{VTBench2025}        & 2025 & 2,933   & -                                  & $\times$ \\
        VTONQA~\cite{wei2024vtonqa}       & 2026 & 8,132   & $256^2 \sim 4096^2$ & $\times$ \\
        \midrule
        \textbf{OpenVTON-Bench (Ours)}    & \textbf{2026} & \textbf{99,925}  & $\mathbf{1024^2 \sim 1536^2}$ & \textbf{$\checkmark$} \\
        \bottomrule
    \end{tabular}
\end{table}

\subsection{Data Sources and Candidate Pool Construction}
\label{app:data_sources}

We construct the initial pool from two primary streams to ensure diversity in fashion styles and imaging conditions.

\vspace{0.5em}
\noindent \textbf{Stream A: Social Media Subset (Refined IG Pair Dataset).}
To capture high-aesthetic ``in-the-wild'' scenarios, we conducted a deep refinement based on the IG Pair Dataset from IMAGDressing-v1~\cite{IMAGDressing-v1}.
\begin{itemize}
    \item \textbf{Manual Cleaning:} We performed strict manual re-screening to enforce a ``one-to-one'' constraint, where one garment image strictly corresponds to one model image, thereby eliminating the redundancy of multiple poses for a single garment.
    \item \textbf{Outcome:} We curated approximately 30,000 high-quality pairs from this source.
\end{itemize}

\vspace{0.5em}
\noindent \textbf{Stream B: Large-scale E-Commerce Subset.}
To ensure categorical coverage, we constructed a corpus from major e-commerce platforms.
\begin{itemize}
    \item \textbf{Collection \& Filtering:} We deployed aesthetic scoring and face detection models during crawling to filter out low-quality or headless images. From over 3 million raw pairs, we retained the high-quality subset.
    \item \textbf{Human Verification:} A team of over 1,000 annotators verified that the standalone garment strictly matches the model's outfit and that no critical body parts are truncated.
\end{itemize}

\vspace{0.5em}
\noindent \textbf{Integration and Resolution Filtering.}
After merging both streams, we applied a strict resolution filter, selecting only images between $1024 \times 1024$ and $1536 \times 1536$. This resulted in a high-quality candidate pool of over 300,000 pairs.

\subsection{Visual Cluster Sampling and Automated Annotation}
\label{app:data_balancing_annotation}

Directly using the candidate pool may lead to long-tail distribution issues. We therefore implemented a feature-driven sampling and annotation strategy.

\vspace{0.5em}
\noindent \textbf{DINOv3 K-Means Sampling.}
We extracted garment embeddings using \textbf{DINOv3}~\cite{dinov3} and applied K-Means with $K=20$ to form visual sampling clusters C0--C19. Final sampling used these clusters as strata, downsampling larger clusters toward an equal quota and retaining source-limited clusters in full. This yielded \textbf{99,925} image pairs with approximately balanced cluster counts, subject to source availability. The 5K test split used throughout the appendix contains \textbf{5,009} pairs. These clusters capture visual variation and are not manually defined garment categories.

\paragraph{Caption-derived Metadata Labels.}
The released \texttt{category} field contains 14 coarse labels obtained by deterministic keyword matching over Gemini captions, using first-match precedence and the merges \texttt{tshirt}$\rightarrow$\texttt{top}, \texttt{knitwear}$\rightarrow$\texttt{sweater}, and \texttt{suit}$\rightarrow$\texttt{jacket}. The labels are blazer, cardigan, coat, dress, hoodie, jacket, jeans, pants, shirt, shorts, skirt, sweater, top, and vest. They are caption-derived metadata, not a manually verified taxonomy. A label can span multiple visual clusters; the final sampling strata are the 20 clusters, not these 14 labels.

The training, validation, and test splits contain 89,927, 4,989, and 5,009 pairs, respectively. Nineteen clusters contain 5,012--5,088 pairs each; C13 contains all 3,340 available pairs from its source cluster. For the complete $14\times20$ label--cluster cross-tabulation, counts by cluster and split, and a comparison of dataset taxonomies, see the \href{https://github.com/RenxingIntelligence/OpenVTON-Bench}{project README}.

\vspace{0.5em}
\noindent \textbf{Cluster Descriptions.}
The raw dataset consists of high-quality \textbf{image pairs}: a \textit{Reference Garment} and a corresponding \textit{Model Image} wearing that garment. To prepare these pairs for standard VTON training and evaluation, we extend them into \textbf{triplets} by generating occlusion masks. We used \textbf{Qwen-VL-Plus}~\cite{Qwen-VL} to suggest garment descriptions from three examples per cluster for downstream mask prompting. These post-hoc descriptions do not define cluster membership or sampling and do not exhaust the variation within a cluster.

\vspace{0.5em}
\noindent \textbf{Automated Triplet Construction.}
We formulate the training triplets $(I_g, I_{gt}, I_m)$, representing the Reference Garment, the Ground-Truth Model, and the Masked Model, respectively, using the following automated pipeline:
\begin{itemize}
    \item \textbf{Segmentation:} Using the garment descriptions as prompts, \textbf{GroundingDINO}~\cite{grounddino} detects the garment regions, and \textbf{SAM3}~\cite{sam3} generates precise pixel-level boundaries.
    \item \textbf{Occlusion (Mask Generation):} Based on the segmentation, we apply a black occlusion layer to the garment region of the original model image. We designate this occluded image as the \textbf{Masked Model} ($I_m$).
\end{itemize}
Consequently, the final dataset contains exactly \textbf{99,925} triplets. During training, the generative model utilizes the pair of the Reference Garment ($I_g$) and the Masked Model ($I_m$) as inputs to reconstruct the target Ground-Truth Model ($I_{gt}$).

\subsection{VLM-assisted Captioning with Gemini-2.0-Flash}
\label{app:captioning}

To enable text-driven editing and fine-grained evaluation, we employed \textbf{Google Gemini-2.0-Flash}~\cite{gemini} for dense captioning. This model was selected for its speed and robust multimodal understanding, which are critical for processing large-scale datasets efficiently.

To ensure high-fidelity descriptions, we implemented a category-aware prompting strategy. We designed two distinct sets of structured prompts, one tailored for upper-body garments and another for lower-body garments. This specialization forces the model to attend to the most relevant attributes for each category, such as sleeve length and neckline for tops versus cut and pattern placement for bottoms, while strictly enforcing definitive language regarding fabric materials. The complete prompts are available in our open-source repository.\setcounter{footnote}{0}\footnote{\url{https://github.com/RenxingIntelligence/OpenVTON-Bench}}

\subsection{Dataset Visualization and Diversity Analysis}
\label{app:data_vis}

\begin{figure}[htbp]
    \centering
    \begin{minipage}[t]{0.49\linewidth}
        \vspace{0pt} 
        \centering
        \begin{minipage}[b]{0.48\linewidth}
            \centering
            \includegraphics[width=\linewidth]{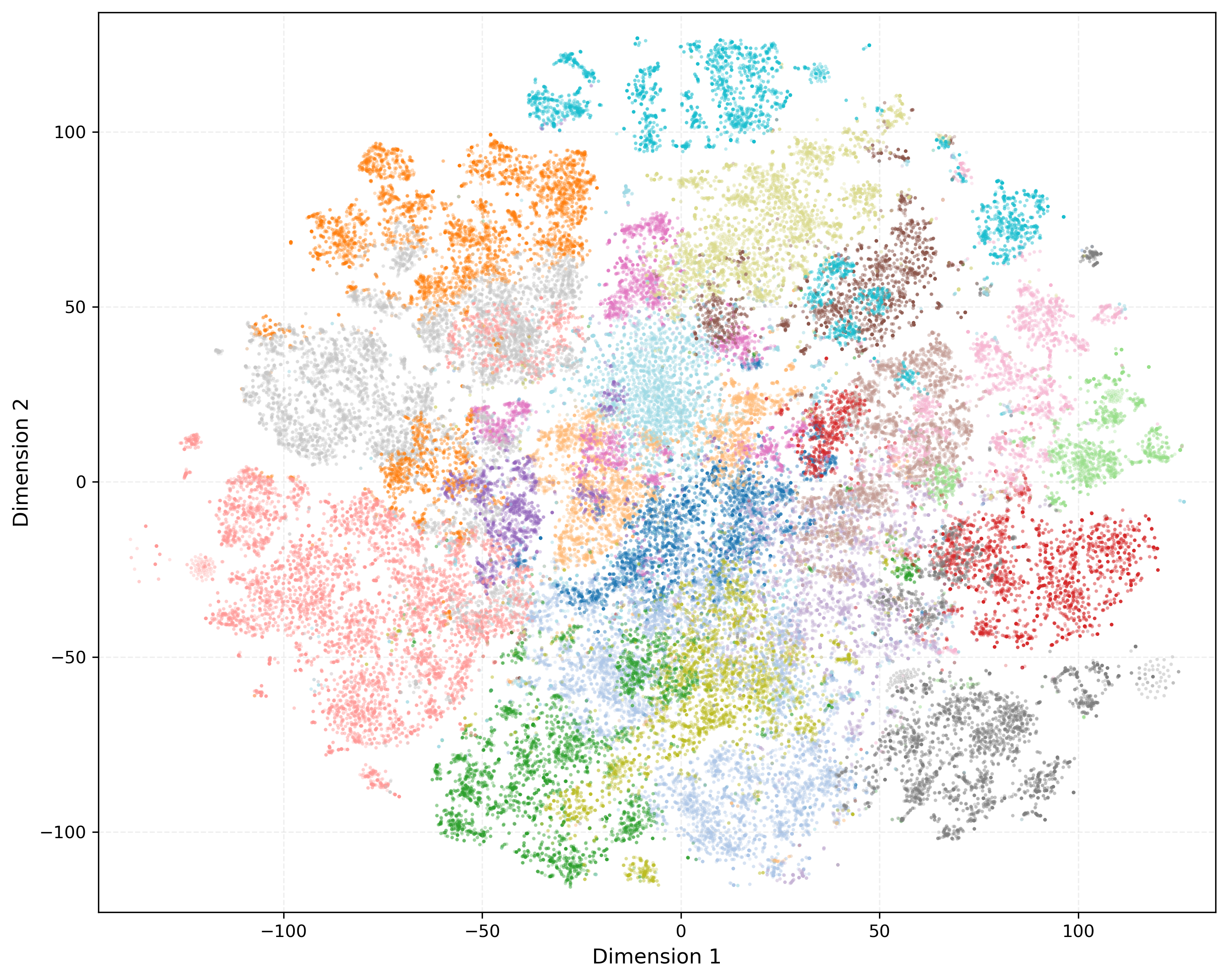}
            \vspace{2pt}
            {\small (a) Full}
        \end{minipage}
        \hfill
        \begin{minipage}[b]{0.48\linewidth}
            \centering
            \includegraphics[width=\linewidth]{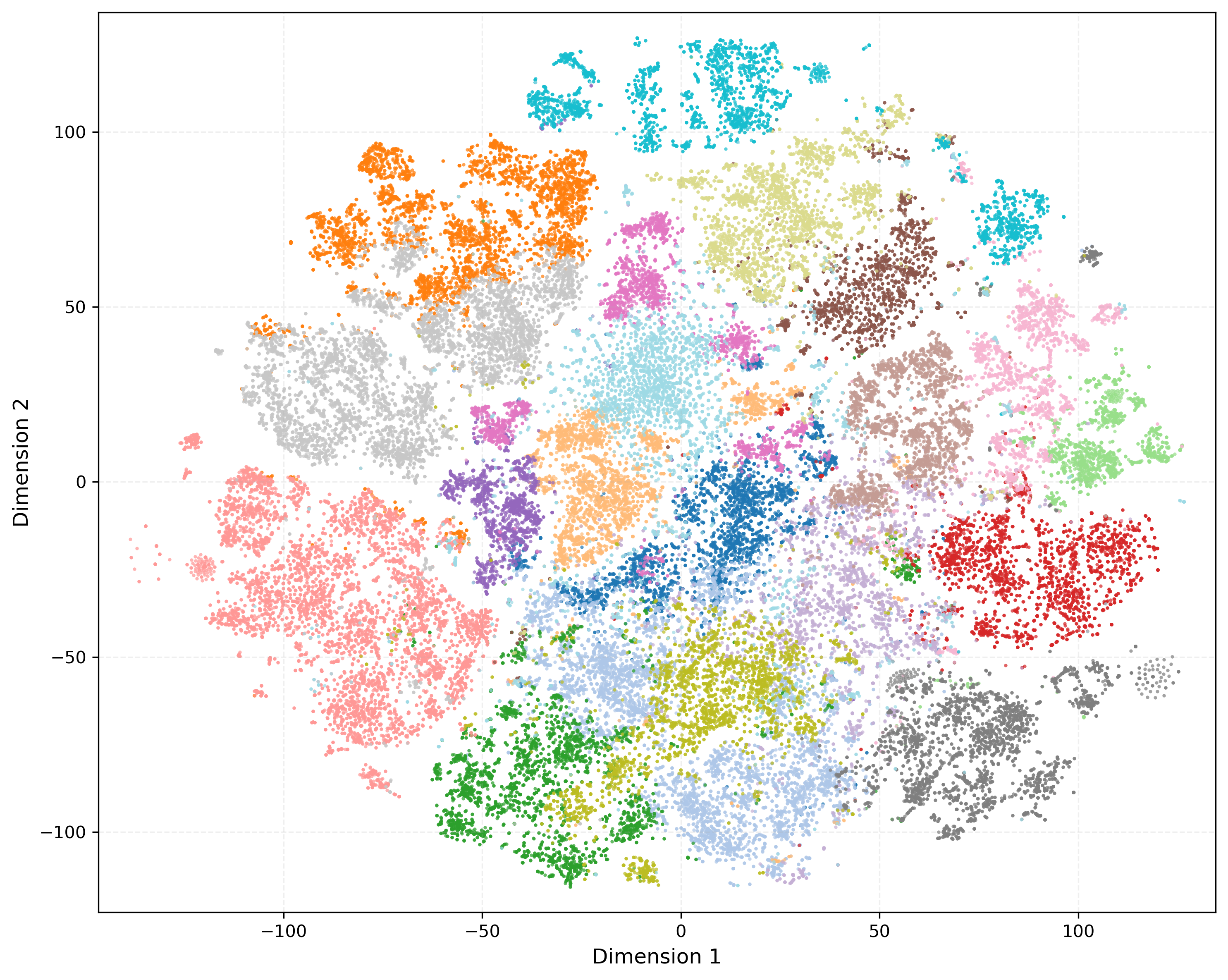}
            \vspace{2pt}
            {\small (b) Train}
        \end{minipage}
        \vspace{0.5em}
        \begin{minipage}[b]{0.48\linewidth}
            \centering
            \includegraphics[width=\linewidth]{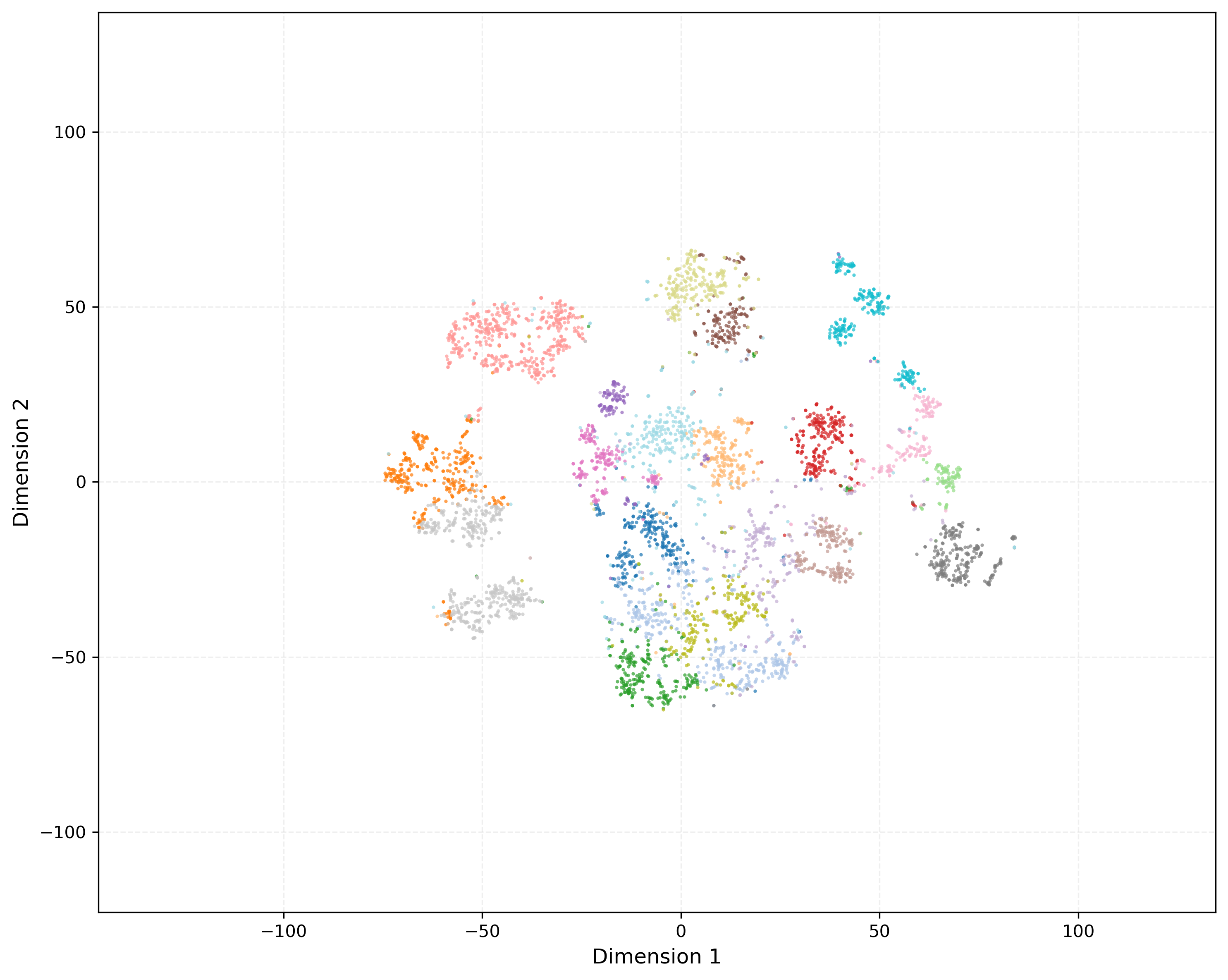}
            \vspace{2pt}
            {\small (c) Validation}
        \end{minipage}
        \hfill
        \begin{minipage}[b]{0.48\linewidth}
            \centering
            \includegraphics[width=\linewidth]{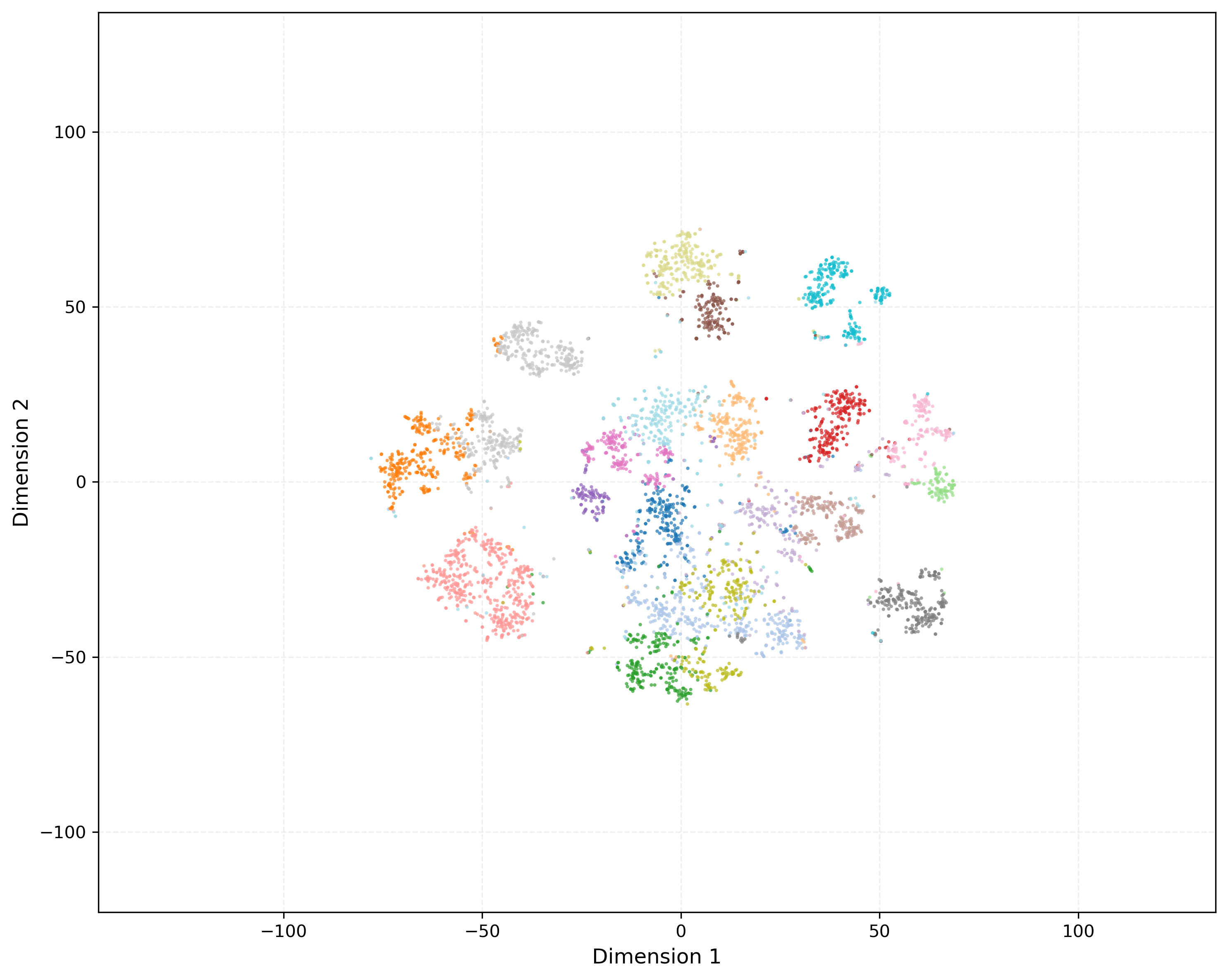}
            \vspace{2pt}
            {\small (d) Test}
        \end{minipage}
    \end{minipage}
    \hfill
    \begin{minipage}[t]{0.49\linewidth}
        \vspace{0pt} 
        \centering
        \includegraphics[width=\linewidth]{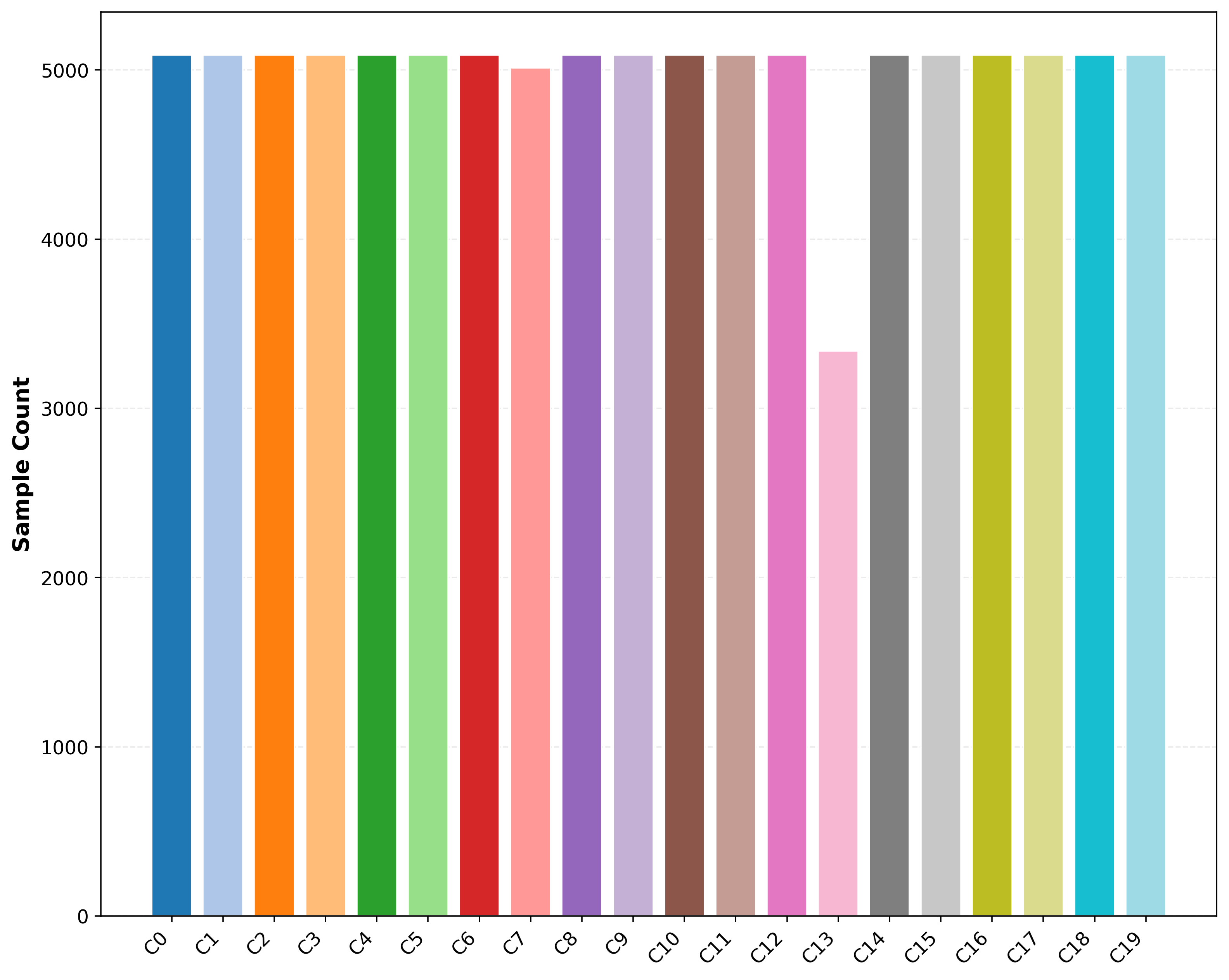}
        \vspace{4pt}
        {\small (e) Visual Cluster Distribution}
    \end{minipage}
    
    \caption{
    \textbf{Dataset Analysis of OpenVTON-Bench.}
    \textbf{Left (a-d):} t-SNE visualizations of the full dataset and the train/validation/test splits.
    \textbf{Right (e):} Counts across the 20 visual sampling clusters, distinct from the 14 caption-derived metadata labels.
    }
    \label{fig:dataset_analysis}
\end{figure}

\begin{figure}[htbp]
    \centering
    \includegraphics[width=\textwidth]{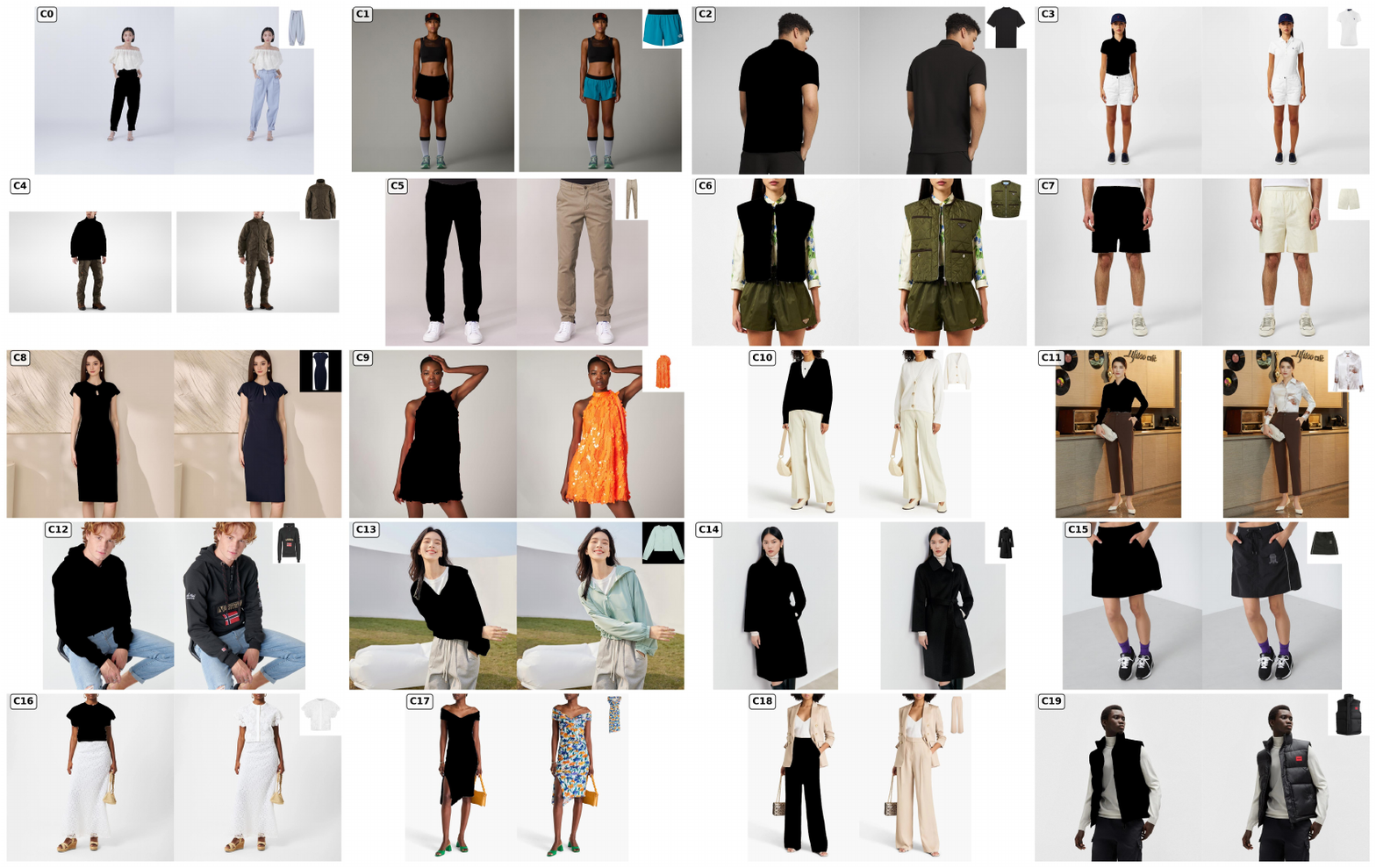}
    \caption{\textbf{Diversity of OpenVTON-Bench.} Example triplets (Garment, Model, Mask) illustrate garment variation across the 20 visual sampling clusters C0--C19. A single example does not exhaust the variation within its cluster.}
    \label{fig:dataset_samples}
\end{figure}

Figure~\ref{fig:dataset_samples} illustrates the visual diversity of OpenVTON-Bench through example triplets. C0--C19 denote visual sampling clusters rather than a fixed garment taxonomy. The examples cover:
\begin{itemize}
    \item \textbf{Tops:} Knit fabrics, different necklines, embroidery, printed patterns, and button closures.
    \item \textbf{Outerwear \& Sleeves:} Varying sleeve lengths, hooded garments, coats, and padded outerwear.
    \item \textbf{Bottoms:} Trousers, shorts, and skirts with different lengths, cuts, and pocket details.
    \item \textbf{Dresses:} Different silhouettes, sleeve designs, and gathered or draped details.
\end{itemize}
These examples illustrate garment variation relevant to VTON evaluation; visual cluster membership is determined by the DINOv3 K-Means assignments.

\FloatBarrier
\section{Evaluation Metrics \& Protocols}
\label{app:metrics}

This section defines the automated evaluation protocol used throughout the main paper and supplementary experiments. We first describe the VLM-as-a-Judge protocol and then give the implementation details of the representation-based metrics.

\subsection{VLM Evaluation Protocol (VLM-as-a-Judge)}
\label{app:vlm_protocol}

Traditional metrics such as FID and SSIM often fail to capture semantic nuances, including local texture preservation and the preservation of non-edited regions. To address this limitation, we propose a VLM-based evaluation protocol using Qwen-VL-Plus as a judge. Unlike standard reference-free evaluations, our protocol performs a “Reference-based Virtual Try-On” assessment by comparing the generated result against both the source garment image and the ground-truth image.

The evaluation covers five dimensions:
\begin{enumerate}
    \item \textit{Background Consistency} ($s_{bg}$): measures whether non-edited regions remain unchanged.
    \item \textit{Person Identity \& Body Consistency} ($s_{id}$): verifies that the model's identity, skin tone, and body structure remain intact.
    \item \textit{Texture Fidelity} ($s_{tex}$): evaluates whether fine garment details, including logos, fabrics, and local patterns, are accurately rendered.
    \item \textit{Shape Preservation} ($s_{shape}$): assesses the geometric correctness of the garment, including sleeve length, neckline, silhouette, and fit.
    \item \textit{Overall Realism} ($s_{real}$): evaluates lighting, shadows, natural folds, and global photographic plausibility.
\end{enumerate}
The aggregate score $s_{avg}$ is the arithmetic mean of these five dimension scores.

\paragraph{Judge Selection.}
We selected Qwen-VL-Plus for multi-image input, structured instruction following, machine-readable scoring, and feasible cost and latency at benchmark scale. This was an engineering choice; the human study validates the selected judge and prompt on the target VTON task. The correlations and disagreement analysis in Appendix~\ref{app:protocol_validation} characterize agreement under the shared rubric.

\paragraph{Evaluation Prompt.}
We designed a structured prompt that forces the VLM to act as a critical fashion image quality evaluator, providing both reasoning and numerical scores for each dimension. The full prompt is available in our open-source repository.

\subsection{Details of Representation-based Metrics}
\label{app:metric_details}

We compute the global similarity $\mathcal{S}_{\text{global}}$ and the multi-scale garment fidelity scores $\mathcal{S}_{\text{rep}}^{(k)}$ using DINOv3~\cite{dinov3} feature representations. The global score $\mathcal{S}_{\text{global}}$ is computed as the cosine similarity between the [CLS] tokens of the generated try-on image $\hat{I}$ and the ground-truth try-on image $I_{gt}$, without mask erosion constraints.

For garment-region fidelity, we apply the generated segmentation mask to the try-on result before feature extraction. Following the main-paper definition, we use a square structural element $B$ of size $3 \times 3$ for morphological erosion and compute four scales:
\begin{itemize}
    \item $\mathbf{k=0}$ ($\mathcal{S}_{\text{rep}}^{(0)}$): uses the full garment mask, capturing both boundary alignment and global texture.
    \item $\mathbf{k=1,2,3}$ ($\mathcal{S}_{\text{rep}}^{(1)}$ to $\mathcal{S}_{\text{rep}}^{(3)}$): use progressively eroded masks, excluding boundary regions as $k$ increases to focus on internal fabric details and reduce penalties caused by minor spatial misalignment.
\end{itemize}
\section{Validation of Evaluation Protocol}
\label{app:protocol_validation}

After defining the automated metrics in Appendix~\ref{app:metrics}, we validate the evaluation protocol from two complementary perspectives: agreement with human model rankings under the shared rubric and agreement among human raters.

\subsection{Correlation Analysis of Automated Metrics}
\label{app:metric_correlation}

We calculated Spearman ($\rho_s$), Kendall ($\rho_k$), and Pearson ($\rho_p$) correlations between automated and human scores aggregated by model. The study collected 92,072 evaluation records from 76 participants, who rated image triplets independently without seeing method identities or VLM outputs, scores, or rationales. Humans and the VLM used the same five dimensions, so these correlations measure agreement under the specified rubric.

Table~\ref{tab:vlm_detail} shows agreement between VLM and human model-level scores, particularly in \textit{Overall Realism} ($\rho_p=0.990$) and \textit{Person Identity \& Body Consistency} ($\rho_p=0.840$). Among the representation metrics in Table~\ref{tab:rep_detail}, the averaged overall score $\bar{\mathcal{S}}$ achieves the highest Spearman correlation ($\rho_s=0.933$). These results support model comparison under this rubric, rather than equivalence to human judgment on every generated image.

\begin{table}[htbp]
    \centering
    \small
    \begin{minipage}[t]{0.48\textwidth}
        \centering
        \caption{\textbf{VLM Metrics per Dimension.} Correlation with human judgments.}
        \label{tab:vlm_detail}
        \begin{tabular}{lccc}
        \toprule
        \textbf{Metric} & $\rho_s \uparrow$ & $\rho_k \uparrow$ & $\rho_p \uparrow$ \\
        \midrule
        $s_{bg}$    & 0.767 & 0.556 & 0.594 \\
        $s_{id}$    & 0.800 & 0.667 & \underline{0.840} \\
        $s_{tex}$   & 0.767 & 0.500 & 0.822 \\
        $s_{shape}$ & \underline{0.867} & \underline{0.722} & 0.806 \\
        $s_{real}$  & \textbf{0.900} & \textbf{0.833} & \textbf{0.990} \\
        $s_{avg}$   & 0.850 & \underline{0.722} & 0.828 \\
        \bottomrule
        \end{tabular}
    \end{minipage}
    \hfill
    \begin{minipage}[t]{0.48\textwidth}
        \centering
        \caption{\textbf{Representation Metrics across Different Dimensions.} Correlation with human judgments.}
        \label{tab:rep_detail}
        \begin{tabular}{lccc}
        \toprule
        \textbf{Metric} & $\rho_s \uparrow$ & $\rho_k \uparrow$ & $\rho_p \uparrow$ \\
        \midrule
        $\mathcal{S}_{\text{global}}$    & \underline{0.867} & \underline{0.722} & 0.687 \\
        $\mathcal{S}_{\text{rep}}^{(0)}$ & 0.850 & 0.667 & \textbf{0.741} \\
        $\mathcal{S}_{\text{rep}}^{(1)}$ & 0.850 & 0.667 & 0.709 \\
        $\mathcal{S}_{\text{rep}}^{(2)}$ & 0.850 & 0.667 & 0.695 \\
        $\mathcal{S}_{\text{rep}}^{(3)}$ & 0.850 & 0.667 & 0.708 \\
        $\bar{\mathcal{S}}_{\text{rep}}$ & 0.850 & 0.667 & \underline{0.712} \\
        $\bar{\mathcal{S}}$              & \textbf{0.933} & \textbf{0.833} & 0.701 \\
        \bottomrule
        \end{tabular}
    \end{minipage}
\end{table}
\FloatBarrier

\subsection{VLM--Human Disagreement}
\label{app:vlm_human_disagreement}

We compare VLM and human mean scores for each of nine models along five dimensions, giving 45 model-by-dimension comparisons. Bias is the mean signed difference (VLM minus human), and MAE is the mean absolute difference on the 1--5 scale. Table~\ref{tab:vlm_human_disagreement} complements the ranking correlations above with score differences. Across the 45 comparisons, MAE is 0.299, median absolute error is 0.202, and 40 differences are within 0.5 points. These are comparisons of aggregate model scores, not image-level prediction errors.

\begin{table}[htbp]
\centering
\small
\caption{\textbf{VLM--human score disagreement.} Bias and MAE are computed across nine model means per dimension. Positive bias indicates higher VLM scores.}
\label{tab:vlm_human_disagreement}
\begin{tabular}{lrr}
\toprule
Dimension & Bias (VLM $-$ human) & MAE \\
\midrule
Background Consistency & $-0.033$ & 0.248 \\
Person Identity \& Body Consistency & $+0.039$ & 0.448 \\
Texture Fidelity & $+0.189$ & 0.299 \\
Shape Preservation & $+0.163$ & 0.305 \\
Overall Realism & $-0.066$ & 0.196 \\
\bottomrule
\end{tabular}
\end{table}

The larger differences are concentrated in UNO, which accounts for four of the five comparisons exceeding 0.5 points. Its VLM scores exceed human scores by 0.947, 1.441, 1.213, and 1.234 for background, identity/body, texture, and shape, respectively. Conversely, mean signed differences for YingHui and NanobananaPro are $-0.236$ and $-0.244$. Thus, agreement in model rankings can coexist with overestimation or underestimation for particular models and dimensions.
\FloatBarrier

\subsection{Detailed Inter-Rater Reliability Analysis}
\label{app:human_consistency}

To rigorously validate the reliability of our human evaluation, we conducted a detailed Inter-Rater Reliability (IRR) analysis. This section focuses on the subset of evaluation tasks that received ratings from at least three independent annotators across all models, which establishes a robust foundation for our statistical analysis.

\subsubsection{Data Preprocessing Pipeline}
We report inter-rater reliability using Krippendorff's Alpha ($\alpha$) after the following preprocessing, applied independently to each dimension:
\begin{itemize}
    \item \textbf{Task-Level Outlier Removal (Denoising):} For each dimension independently, we compute the median score within each task. Rating records deviating from this task-specific median by more than a threshold $\tau$ are discarded as outliers.
    \item \textbf{User-Level Mean-Centering (Alignment):} For the surviving records, we perform mean-centering grouped by the annotator's ID. This adjusts for differences in the mean scores assigned by stricter or more lenient raters.
    \item \textbf{Composite Score Aggregation:} While the five dimensions are processed independently, the Composite Score is analyzed using two strategies: (1) Aggregated from Filtered, where a record is averaged only if it survives the outlier removal across all five dimensions; (2) Filtered Directly, where the raw average of the five dimensions is computed and then passed through the denoising and alignment pipeline as an independent metric.
\end{itemize}

\subsubsection{Main Protocol: Threshold $\tau = 1.0$}
Our primary evaluation protocol uses an outlier removal threshold of $\tau = 1.0$. Relative to the stricter threshold below, this setting retains more rating records while yielding composite agreement of approximately 0.70.

As shown in Table~\ref{tab:irr_threshold_1}, agreement after preprocessing varies by dimension, from $\alpha=0.468$ for background to $0.835$ for realism. The two strategies for calculating the Composite Average yield similar agreement scores ($\alpha = 0.700$ vs. $0.706$).

\begin{table}[htbp]
\centering
\caption{\textbf{Inter-Rater Reliability (Threshold $\tau = 1.0$).} Results after task-level outlier removal and user-level mean-centering. CI denotes the 95\% confidence interval.}
\label{tab:irr_threshold_1}
\begin{adjustbox}{width=\linewidth}
\begin{tabular}{lcccc}
\toprule
\textbf{Metric / Dimension} & \textbf{Proposed $\alpha$} & \textbf{95\% CI} & \textbf{Tasks Retained} & \textbf{Records Retained} \\
\midrule
Background Consistency & 0.468 &[0.124, 0.610] & 100.0\% & 99.0\% \\
Person Consistency & 0.680 &[0.624, 0.730] & 99.7\% & 95.6\% \\
Texture Fidelity & 0.518 &[0.479, 0.557] & 99.9\% & 94.4\% \\
Shape Preservation & 0.748 &[0.694, 0.791] & 99.7\% & 96.1\% \\
Overall Realism & 0.835 &[0.800, 0.870] & 99.8\% & 95.9\% \\
\midrule
Composite (Aggregated from Filtered) & 0.700 &[0.613, 0.757] & 96.2\% & 87.1\% \\
Composite (Filtered Directly) & 0.706 &[0.646, 0.741] & 99.7\% & 97.3\% \\
\bottomrule
\end{tabular}
\end{adjustbox}
\end{table}

\subsubsection{Sensitivity Analysis: Stricter Threshold $\tau = 0.8$}
To further analyze the robustness of our evaluation, we conducted a sensitivity analysis using a stricter threshold of $\tau = 0.8$. This setting more aggressively penalizes deviations from the task median.

As presented in Table~\ref{tab:irr_threshold_08}, a stricter threshold increases the individual $\alpha$ scores, with Shape Preservation reaching $0.922$, while retaining fewer records. For the \textit{Composite (Aggregated from Filtered)} strategy, retaining a record only when it passes filtering in all five dimensions reduces retention to 50.1\%. We use $\tau = 1.0$ to preserve broader coverage; the higher agreement at $\tau = 0.8$ should be interpreted alongside this reduction in retained data.

\begin{table}[htbp]
\centering
\caption{\textbf{Sensitivity Analysis of IRR (Threshold $\tau = 0.8$).} A stricter threshold further improves $\alpha$ scores but significantly reduces the percentage of retained records.}
\label{tab:irr_threshold_08}
\begin{adjustbox}{width=\linewidth}
\begin{tabular}{lcccc}
\toprule
\textbf{Metric / Dimension} & \textbf{Proposed $\alpha$} & \textbf{95\% CI} & \textbf{Tasks Retained} & \textbf{Records Retained} \\
\midrule
Background Consistency & 0.801 &[0.338, 0.891] & 99.8\% & 95.8\% \\
Person Consistency & 0.866 &[0.818, 0.891] & 97.2\% & 81.5\% \\
Texture Fidelity & 0.697 & [0.664, 0.726] & 97.2\% & 73.9\% \\
Shape Preservation & 0.922 &[0.903, 0.936] & 98.2\% & 85.0\% \\
Overall Realism & 0.923 &[0.872, 0.943] & 98.7\% & 90.2\% \\
\midrule
Composite (Aggregated from Filtered) & 0.710 &[0.475, 0.794] & 77.3\% & 50.1\% \\
Composite (Filtered Directly) & 0.712 &[0.650, 0.758] & 99.5\% & 96.3\% \\
\bottomrule
\end{tabular}
\end{adjustbox}
\end{table}

\vspace{0.5em}
\noindent \textbf{Summary.}
The reported agreement characterizes ratings after outlier removal and user-level mean-centering. Agreement is stronger for realism and shape than for background and texture. The threshold comparison shows the trade-off between agreement and data retention; it does not establish that the raw ratings are free of individual biases.
\FloatBarrier

\subsection{Color Consistency Analysis}
\label{app:color_consistency}

Color contributes to the five-dimensional VLM rubric but is not reported as a separate score. We therefore measure color differences between generated test outputs and their paired ground-truth images using CIEDE2000~\cite{luo2001ciede2000}. This diagnostic supplements the original metrics without changing their scores or aggregation.

\paragraph{Protocol.}
We resize RGB images to $256\times256$ with LANCZOS interpolation and target garment masks with nearest-neighbor interpolation. RGB values are converted to CIELAB using the D65 reference white. To reduce boundary effects, we erode the garment mask by five pixels and exclude its five-pixel dilation from the non-edited region. We use the same valid sample IDs across all methods, requiring readable outputs and at least 64 pixels in each region after mask processing. For each image, region-mean $\Delta E_{00}$ compares the mean Lab colors within the corresponding regions, measuring overall color cast. Pixel-mean $\Delta E_{00}$ averages the differences at corresponding garment pixels and is also sensitive to texture and spatial misalignment. Table~\ref{tab:color_consistency} reports the mean of these per-image measurements; lower values indicate smaller color differences.

\begin{table}[htbp]
\centering
\small
\caption{\textbf{Color consistency on generated test outputs.} CIEDE2000 color differences are evaluated under a common protocol. Best values in each column are bold.}
\label{tab:color_consistency}
\begin{tabular}{lccc}
\toprule
\multirow{2}{*}{Method} & \multicolumn{2}{c}{Garment $\Delta E_{00}\downarrow$} & Non-edited $\Delta E_{00}\downarrow$ \\
\cmidrule(lr){2-3}
 & Region-mean & Pixel-mean & Region-mean \\
\midrule
OOTD & 12.67 & 16.85 & 1.57 \\
UNO & 12.84 & 20.08 & 10.50 \\
EasyControl & 12.70 & 17.31 & 2.44 \\
FLUX.1-Kontext-dev & 10.01 & 15.19 & 4.85 \\
FLUX.2-dev & 5.21 & 9.38 & 1.91 \\
NanobananaPro & 6.10 & 11.57 & 2.18 \\
Qwen-Editor & 6.21 & 11.61 & 2.18 \\
HuiWa & \textbf{4.46} & 8.98 & \textbf{0.30} \\
YingHui & 4.71 & \textbf{8.45} & 1.02 \\
\bottomrule
\end{tabular}
\end{table}

\paragraph{Results.}
YingHui has the lowest garment pixel-mean error and the second-lowest region-mean errors for both garment and non-edited regions. HuiWa has the lowest region-mean errors in both regions. The local color shift in Figure~\ref{fig:qualitative_more} can therefore coexist with strong average color preservation. Reporting color differences separately complements the five-dimensional rubric, whose aggregate can obscure a localized chromatic error.
\FloatBarrier
\section{Additional Experimental Results}
\label{app:results}

This section provides supplementary implementation details and experimental results that were omitted from the main text due to space constraints. We first report the hardware environment and baseline configurations, then evaluate additional baselines, analyze the effectiveness of fine-tuning on OpenVTON-Bench, and finally provide extended qualitative comparisons.

\subsection{Implementation Details and Experimental Setup}
\label{app:exp_setup}

\subsubsection{Hardware and Software Environment}
All evaluations, including both main-paper and supplementary baselines, as well as the fine-tuning experiments in Appendix~\ref{app:finetuning_experiments}, were conducted on a high-performance computing cluster to ensure consistent benchmarking.
\begin{itemize}
    \item \textbf{GPU:} NVIDIA A800 (80GB VRAM) $\times$ 8.
    \item \textbf{Frameworks:} PyTorch 2.9.1, CUDA 12.4.
    \item \textbf{Precision:} Mixed precision (BF16 or Float32) was used during inference to align with standard deployment environments.
\end{itemize}

\subsubsection{Baseline Configurations}
Public-checkpoint baselines use official released weights and configurations. FLUX.1-Kontext-dev uses image-editing weights, and FLUX.2-dev uses native multi-reference editing. API-based results refer to the fixed generated outputs used in this evaluation. For the model identifiers, checkpoints, native image settings, sampling steps, guidance scales, and seeds of all main-paper and supplementary baselines, see the \href{https://github.com/RenxingIntelligence/OpenVTON-Bench}{project README}.

Before the main metric computation, source, ground-truth, and generated images undergo the same aspect-ratio-preserving LANCZOS resizing and centered white padding to a $1024\times1024$ canvas. This common evaluation size is separate from each model's native inference size. The additional color diagnostic uses the $256\times256$ analysis protocol described in Appendix~\ref{app:color_consistency}.

\subsection{Evaluation of Supplementary Baselines}
\label{app:more_baselines}

To provide a more comprehensive overview of existing virtual try-on methodologies, we conduct supplementary evaluations on three additional well-known baselines: \textbf{DCI-VTON}~\cite{gou2023taming}, \textbf{PromptDresser}~\cite{PromptDresser2025}, and \textbf{StableVITON}~\cite{stableviton2024}. This evaluation includes both our proposed VLM-based semantic scoring and standard objective metrics, evaluated on the 5K test set containing \textbf{5,009} pairs.

This supplementary experiment reveals a clear divergence between objective pixel-level reconstruction metrics and semantic/distributional realism, which supports the core motivation of OpenVTON-Bench: traditional metrics do not fully reflect human perceptual quality.

\textbf{Semantic and Perceptual Quality (VLM \& FID).}
As reported in Table~\ref{tab:more_baselines_vlm}, \textbf{PromptDresser} achieves the highest scores across all VLM semantic dimensions among the three supplementary baselines, with an overall score of $s_{avg}=4.095$. Its results in \textit{Texture Fidelity} and \textit{Shape Preservation} indicate stronger performance under the specified VLM rubric. Its Fr\'echet Inception Distance (FID) score of 32.31 (Table~\ref{tab:more_baselines_obj}) is also the best among the three, indicating that its generated image distribution is closer to the reference distribution.

\textbf{Pixel-level and Structural Fidelity.}
Conversely, when evaluating strict point-to-point reconstruction (Table~\ref{tab:more_baselines_obj}), \textbf{StableVITON} dominates almost all pixel-alignment (SSIM, PSNR) and perceptual (LPIPS, representation similarity) metrics in both the isolated garment area and the global image. For instance, its global SSIM (0.858) and PSNR (20.74 dB) significantly outperform the others.

This contrast, where StableVITON excels in pixel-wise reconstruction while PromptDresser performs best in semantic realism and distribution quality, highlights the necessity of the hybrid evaluation protocol proposed in OpenVTON-Bench. Relying solely on SSIM/PSNR would penalize the realistic but structurally varied outputs of PromptDresser, whereas using only FID ignores StableVITON's stronger condition adherence. DCI-VTON achieves intermediate objective scores but struggles with both VLM scores and FID (51.60), indicating limitations in overall photorealism when faced with the in-the-wild complexities of our benchmark.

\begin{table}[htbp]
\centering
\caption{\textbf{Supplementary Baseline Evaluation: VLM Semantic Scoring.}
Scores (1--5 scale) across five semantic dimensions. \textbf{Bold} indicates the best result, and \underline{underline} indicates the second best.}
\label{tab:more_baselines_vlm}
\begin{adjustbox}{width=0.9\textwidth}
\begin{tabular}{lcccccc}
\toprule
\textbf{Method}
& \textbf{$s_{bg} \uparrow$}
& \textbf{$s_{id} \uparrow$}
& \textbf{$s_{tex} \uparrow$}
& \textbf{$s_{shape} \uparrow$}
& \textbf{$s_{real} \uparrow$}
& \textbf{$s_{avg} \uparrow$} \\
\midrule
DCI-VTON~\cite{gou2023taming} & 4.029 & 3.597 & 3.255 & 3.674 & 3.773 & 3.665 \\
PromptDresser~\cite{PromptDresser2025} & \textbf{4.428} & \textbf{3.942} & \textbf{3.693} & \textbf{4.221} & \textbf{4.193} & \textbf{4.095} \\
StableVITON~\cite{stableviton2024} & \underline{4.296} & \underline{3.846} & \underline{3.259} & \underline{3.905} & \underline{3.914} & \underline{3.844} \\
\bottomrule
\end{tabular}
\end{adjustbox}
\end{table}

\begin{table}[htbp]
\centering
\caption{\textbf{Supplementary Baseline Evaluation: Objective Metrics.}
Evaluation is divided into representation-based and pixel-based groups.}
\label{tab:more_baselines_obj}
\begin{adjustbox}{width=\linewidth}
\begin{tabular}{l ccc cccc}
\toprule
\multirow{2}{*}{\textbf{Method}} & \multicolumn{3}{c}{\textbf{Representation}} & \multicolumn{4}{c}{\textbf{Pixel}} \\
\cmidrule(lr){2-4} \cmidrule(lr){5-8}
& $\mathcal{S}_{\text{global}} \uparrow$ & $\bar{\mathcal{S}}_{\text{rep}} \uparrow$ & $\bar{\mathcal{S}} \uparrow$ & \textbf{PSNR} $\uparrow$ & \textbf{SSIM} $\uparrow$ & \textbf{LPIPS} $\downarrow$ & \textbf{FID} $\downarrow$ \\
\midrule
DCI-VTON~\cite{gou2023taming} & 0.836 & 0.724 & 0.780 & \underline{16.303} & \underline{0.812} & \underline{0.256} & 51.599 \\
PromptDresser~\cite{PromptDresser2025} & \underline{0.871} & \underline{0.756} & \underline{0.814} & 16.273 & 0.809 & 0.257 & \textbf{32.307} \\
StableVITON~\cite{stableviton2024} & \textbf{0.901} & \textbf{0.788} & \textbf{0.845} & \textbf{20.742} & \textbf{0.858} & \textbf{0.134} & \underline{34.559} \\
\bottomrule
\end{tabular}
\end{adjustbox}
\end{table}
\FloatBarrier

\subsection{Effectiveness of Fine-Tuning on OpenVTON-Bench}
\label{app:finetuning_experiments}

To evaluate OpenVTON-Bench as a training resource, we fine-tuned \textbf{DCI-VTON} and \textbf{StableVITON} on our training set for 1,000 steps and evaluated them on the 5K test set (5,009 pairs). Table~\ref{tab:finetuning_merged} shows improved FID and representation-level similarity for both models relative to their original VITON-HD checkpoints. FID decreases from 59.145 to \textbf{19.955} for DCI-VTON and from 35.630 to \textbf{25.738} for StableVITON. StableVITON's garment representation score ($\bar{\mathcal{S}}_{\text{rep}}$) increases from 0.722 to 0.737, and its VLM score ($s_{avg}$) reaches 3.850. The gains vary by metric: DCI-VTON's VLM score decreases from 3.511 to 3.473, while StableVITON's SSIM changes from 0.817 to 0.816.

These results support the value of OpenVTON-Bench for adapting existing models to its broader image distribution. The improvements in FID and garment representation similarity complement the evaluation results, while the differing changes across metrics reinforce the need for a multi-dimensional assessment.

\begin{table}[htbp]
\centering
\caption{\textbf{Fine-Tuning Results on OpenVTON-Bench (Evaluated on the 5K Test Set).}
Both models improve in FID and representation similarity after fine-tuning. \textbf{Bold} indicates the better result within each model group.}
\label{tab:finetuning_merged}
\begin{adjustbox}{width=\linewidth}
\begin{tabular}{l c c ccc cccc}
\toprule
\multirow{2}{*}{\textbf{Method}} & \multirow{2}{*}{\textbf{Training Data}} & \textbf{Semantic} & \multicolumn{3}{c}{\textbf{Representation}} & \multicolumn{4}{c}{\textbf{Pixel}} \\
\cmidrule(lr){3-3} \cmidrule(lr){4-6} \cmidrule(lr){7-10}
& & \textbf{$s_{avg}$ $\uparrow$} & $\mathcal{S}_{\text{global}} \uparrow$ & $\bar{\mathcal{S}}_{\text{rep}} \uparrow$ & $\bar{\mathcal{S}} \uparrow$ & \textbf{PSNR} $\uparrow$ & \textbf{SSIM} $\uparrow$ & \textbf{LPIPS} $\downarrow$ & \textbf{FID} $\downarrow$ \\
\midrule
\multirow{2}{*}{DCI-VTON} & VITON-HD Train & \textbf{3.511} & 0.829 & 0.657 & 0.743 & 15.686 & 0.803 & 0.292 & 59.145 \\
& OpenVTON Train (Ours) & 3.473 & \textbf{0.858} & \textbf{0.675} & \textbf{0.767} & \textbf{19.610} & \textbf{0.863} & \textbf{0.195} & \textbf{19.955} \\
\midrule
\multirow{2}{*}{StableVITON} & VITON-HD Train & 3.770 & 0.854 & 0.722 & 0.788 & 17.519 & \textbf{0.817} & 0.274 & 35.630 \\
& OpenVTON Train (Ours) & \textbf{3.850} & \textbf{0.870} & \textbf{0.737} & \textbf{0.804} & \textbf{17.862} & 0.816 & \textbf{0.271} & \textbf{25.738} \\
\bottomrule
\end{tabular}
\end{adjustbox}
\end{table}
\FloatBarrier

\subsection{Extended Qualitative Comparison}
\label{app:extended_qualitative}

\begin{figure}[htbp]
    \centering
    \includegraphics[width=\textwidth]{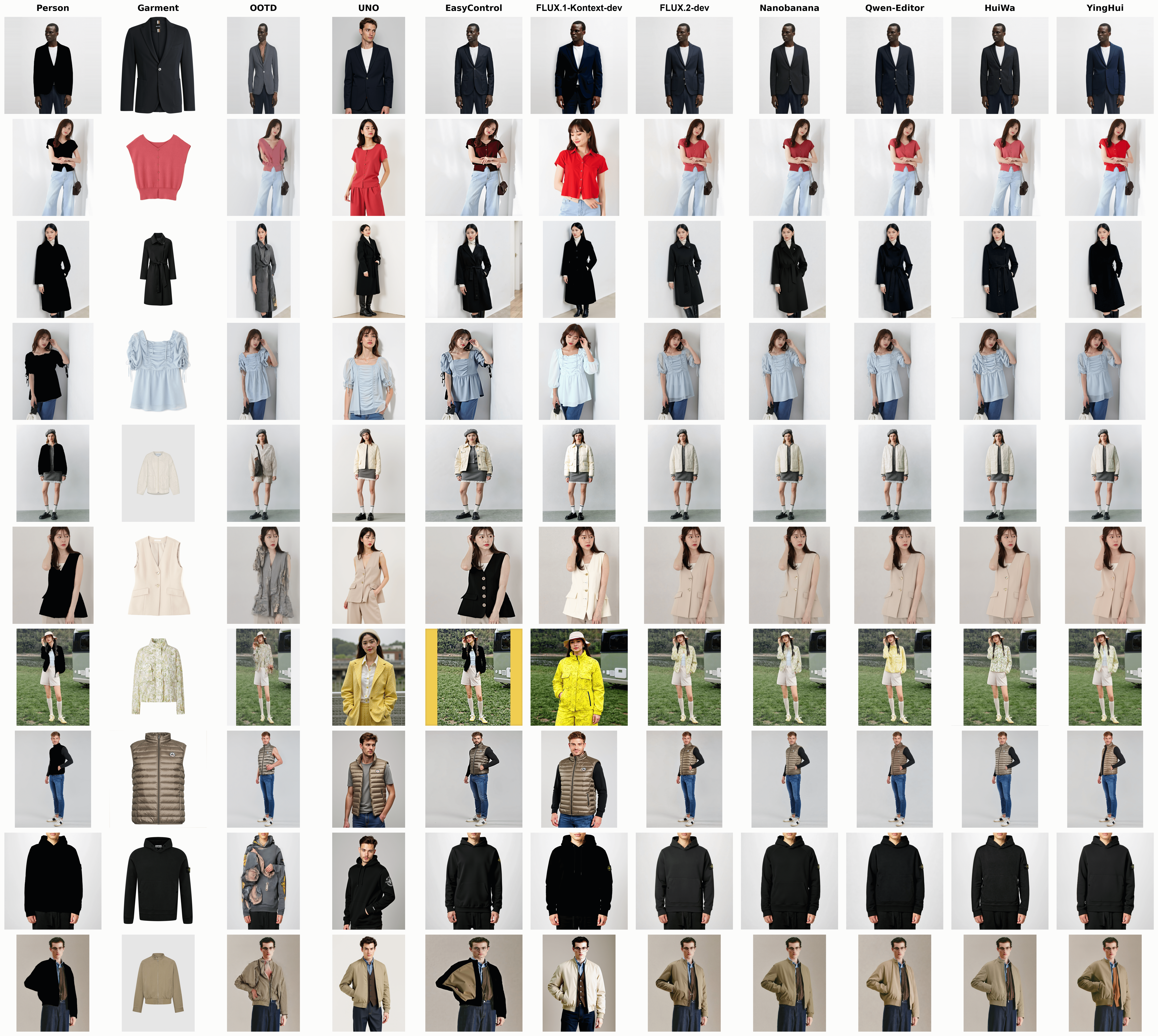}
    \caption{\textbf{Extended qualitative comparison.} We compare the generation results of different baseline models on challenging OpenVTON-Bench examples.}
    \label{fig:qualitative_more}
\end{figure}

We provide a comprehensive visual comparison of all benchmarked methods. Figure~\ref{fig:qualitative_more} illustrates generation results for different models evaluated in the main paper and supplementary experiments.
For each model, we present \textbf{10 generated samples} corresponding to challenging input scenarios from OpenVTON-Bench. This allows a holistic assessment of each model's ability to handle intricate details, such as text and logos, while maintaining garment integrity under complex poses.

%
\section{Ethics, Licenses, Human Study, and Compute}
\label{app:ethics_compute}

\subsection{Broader Impacts and Safeguards}

OpenVTON-Bench is intended to support reliable and transparent evaluation of controllable virtual try-on systems. Its positive impact lies in helping the community identify failures in garment fidelity, texture preservation, body-shape consistency, and background preservation, thereby reducing misleading claims based on coarse image metrics.

At the same time, virtual try-on technology may be misused for deceptive synthetic fashion imagery, unauthorized person-image editing, or unrealistic body presentation. OpenVTON-Bench is intended for non-commercial research and evaluation. We advise against using it for biometric identification, face recognition, surveillance, or attempts to recover private identity information. Facial regions are anonymized before release, unnecessary metadata is removed, and a takedown mechanism is provided for individuals or copyright holders.

\subsection{Licenses and Existing Assets}

The \href{https://huggingface.co/datasets/RenxingIntelligence/OpenVTON}{public dataset} is released under Creative Commons Attribution--NonCommercial 4.0 International (CC BY-NC 4.0). This license permits sharing and adaptation for non-commercial purposes subject to its attribution and other conditions. Original asset terms and third-party rights continue to apply. Table~\ref{tab:licenses} summarizes the main external assets used in this work; model and API use is governed by the corresponding provider terms.

\begin{table}[t]
\centering
\small
\caption{Main external assets used in OpenVTON-Bench.}
\label{tab:licenses}
\begin{tabular}{p{0.22\linewidth} p{0.34\linewidth} p{0.32\linewidth}}
\toprule
Asset & Role & License / Terms \\
\midrule

IMAGDressing-v1 / IGPair~\cite{IMAGDressing-v1} 
& Source of curated VTON pairs 
& Original dataset terms apply; academic / personal use only; commercial use prohibited \\

Public online images 
& Web-scale candidate pool 
& Benchmark release: CC BY-NC 4.0; original rights apply; takedown supported \\

DINOv3~\cite{dinov3} 
& Semantic clustering and representation metric 
& Meta DINOv3 License / Agreement \\

Gemini 2.0 Flash~\cite{gemini} 
& Dense garment captioning 
& Google API / model terms apply \\

Qwen-VL-Plus~\cite{Qwen-VL} 
& VLM-as-a-Judge evaluation 
& Qwen API / model terms apply \\

SAM3~\cite{sam3} 
& Garment mask generation and segmentation 
& Meta SAM License \\

GroundingDINO~\cite{grounddino} 
& Open-vocabulary garment localization 
& Apache-2.0 license \\

\bottomrule
\end{tabular}
\end{table}

\subsection{Human Evaluation Protocol}

Our human perceptual study was conducted only for image-quality assessment. Participants independently evaluated randomly sampled VTON triplets on a five-point Likert scale along the same five dimensions as the VLM: background consistency, person identity and body consistency, texture fidelity, shape preservation, and overall realism. They did not see method identities or any VLM outputs, scores, or rationales. They were instructed to judge try-on quality only, and not to infer personal identity, demographic attributes, or any sensitive information about depicted individuals.

Participation was voluntary. Participants were informed of the purpose of the study, the expected task, and the use of aggregated anonymous ratings for research evaluation. No personally identifying information about participants is released. All participants involved in annotation or perceptual evaluation were compensated according to local labor requirements and platform norms, with an estimated hourly rate no lower than the applicable local minimum wage.

\subsection{IRB or Equivalent Ethical Review}
The human study involved anonymous perceptual ratings of images and did not collect sensitive personal information from participants. No formal IRB approval or equivalent review was obtained. We followed the practices described above, including informed participation, voluntary withdrawal, fair compensation, anonymized aggregation of responses, and no release of participant-identifying information.

\newpage
\subsection{Compute Resources}

All evaluations were conducted on a high-performance computing cluster with NVIDIA A800 GPUs, PyTorch 2.9.1, CUDA 12.4, and BF16 or Float32 inference. Each benchmarked method was evaluated using official checkpoints, official APIs, or provider-recommended inference settings whenever available.

For reproducibility, we report the compute configuration used for the main experiments. All experiments were conducted on a high-performance cluster equipped with NVIDIA A800 GPUs, each with 80GB VRAM. For open-source local models, inference on the full 5K-image test split was performed on a single A800 GPU and typically required 4--10 hours of wall-clock time, depending on the model architecture, output resolution, batching strategy, and implementation efficiency. Thus, the main evaluation cost for each open-source model is approximately 4--10 GPU-hours. API-based commercial systems were evaluated through their official services, so their backend compute cost is not directly observable. Additional exploratory runs were used for data filtering, prompt validation, and metric validation; however, these runs are not required to reproduce the reported benchmark results.
\clearpage
\section*{NeurIPS Paper Checklist}
\begin{enumerate}

\item {\bf Claims}
    \item[] Question: Do the main claims made in the abstract and introduction accurately reflect the paper's contributions and scope?
    \item[] Answer:\answerYes{}
    \item[] Justification: Sections 3--5 support the dataset and hybrid evaluation contributions. The paper distinguishes 20 DINOv3 K-Means sampling clusters from 14 caption-derived labels and limits human validation to model-level agreement under the shared rubric. Appendix~\ref{app:protocol_validation} provides correlation, disagreement, reliability, and color analyses.

\item {\bf Limitations}
    \item[] Question: Does the paper discuss the limitations of the work performed by the authors?
    \item[] Answer: \answerYes{}
    \item[] Justification: Section 6 discusses foundation-model biases, under-representation of extreme poses and occlusions, and the distinction between 2D visual evaluation and physical fit, sizing, grading, or material physics. Appendix~\ref{app:protocol_validation} reports disagreement and the dependence of inter-rater reliability on preprocessing.

\item {\bf Theory assumptions and proofs}
    \item[] Question: For each theoretical result, does the paper provide the full set of assumptions and a complete (and correct) proof?
    \item[] Answer: \answerNA{} 
    \item[] Justification: The paper does not present formal theoretical theorems or proofs. The equations in Section 4 define the evaluation protocol and metrics rather than theoretical results requiring proof.

    \item {\bf Experimental result reproducibility}
    \item[] Question: Does the paper fully disclose all the information needed to reproduce the main experimental results of the paper to the extent that it affects the main claims and/or conclusions of the paper (regardless of whether the code and data are provided or not)?
    \item[] Answer: \answerYes{} 
    \item[] Justification: Sections 3--5 and Appendices~\ref{app:metrics} and~\ref{app:exp_setup} specify the sampling and evaluation protocols, aggregation, preprocessing, and baseline setup, with detailed configurations documented in the project README. Public-checkpoint generation is distinguished from evaluation of fixed API outputs. The public repository provides evaluation code and prompts.

\item {\bf Open access to data and code}
    \item[] Question: Does the paper provide open access to the data and code, with sufficient instructions to faithfully reproduce the main experimental results, as described in supplemental material?
    \item[] Answer: \answerYes{} 
    \item[] Justification: The abstract links to the public evaluation code, whose README provides installation and execution instructions and links to the Hugging Face dataset. The dataset is released under CC BY-NC 4.0. Appendix~\ref{app:exp_setup} distinguishes model inference settings from common evaluation preprocessing and identifies fixed-output evaluation for API-based systems.

\item {\bf Experimental setting/details}
    \item[] Question: Does the paper specify all the training and test details (e.g., data splits, hyperparameters, how they were chosen, type of optimizer) necessary to understand the results?
    \item[] Answer: \answerYes{} 
    \item[] Justification: Appendix~\ref{app:data_balancing_annotation} gives exact split sizes. Appendices~\ref{app:metrics}, \ref{app:color_consistency}, and~\ref{app:exp_setup} describe metric definitions, the color diagnostic, hardware, and common preprocessing. Detailed baseline configurations are documented in the project README; Appendix~\ref{app:finetuning_experiments} describes the supplementary fine-tuning comparison.

\item {\bf Experiment statistical significance}
    \item[] Question: Does the paper report error bars suitably and correctly defined or other appropriate information about the statistical significance of the experiments?
    \item[] Answer: \answerYes{} 
    \item[] Justification: Appendix~\ref{app:human_consistency} reports 95\% confidence intervals for inter-rater reliability after the specified preprocessing and compares two filtering thresholds. The model-level correlations and disagreement statistics are descriptive. We do not report repeated-run error bars for all generated outputs because of the cost of full high-resolution inference.

\item {\bf Experiments compute resources}
    \item[] Question: For each experiment, does the paper provide sufficient information on the computer resources (type of compute workers, memory, time of execution) needed to reproduce the experiments?
    \item[] Answer: \answerYes{} 
    \item[] Justification: Appendix~\ref{app:ethics_compute} reports the compute configuration used in the experiments, including the NVIDIA A800 GPU environment, 80GB VRAM, software stack, inference precision, and single-GPU inference cost.
    
\item {\bf Code of ethics}
    \item[] Question: Does the research conducted in the paper conform, in every respect, with the NeurIPS Code of Ethics \url{https://neurips.cc/public/EthicsGuidelines}?
    \item[] Answer: \answerYes{} 
    \item[] Justification: Appendix~\ref{app:ethics_compute} describes facial anonymization, CC BY-NC 4.0 licensing, a takedown mechanism, informed participation, and compensation. The benchmark is intended for responsible non-commercial research and evaluation; the absence of formal IRB approval is stated explicitly.

\item {\bf Broader impacts}
    \item[] Question: Does the paper discuss both potential positive societal impacts and negative societal impacts of the work performed?
    \item[] Answer: \answerYes{} 
    \item[] Justification: Appendix~\ref{app:ethics_compute} discusses more transparent VTON evaluation and potential misuse, including deceptive imagery, unauthorized person-image editing, and unrealistic body presentation. It describes facial anonymization, non-commercial licensing, responsible-use recommendations, and takedown support.
    
\item {\bf Safeguards}
    \item[] Question: Does the paper describe safeguards that have been put in place for responsible release of data or models that have a high risk for misuse (e.g., pre-trained language models, image generators, or scraped datasets)?
    \item[] Answer: \answerYes{} 
    \item[] Justification: Appendix~\ref{app:ethics_compute} describes facial anonymization, metadata removal, a takedown mechanism, and CC BY-NC 4.0 licensing. Recommendations against identity inference and surveillance are stated separately from the license terms.

\item {\bf Licenses for existing assets}
    \item[] Question: Are the creators or original owners of assets (e.g., code, data, models), used in the paper, properly credited and are the license and terms of use explicitly mentioned and properly respected?
    \item[] Answer: \answerYes{} 
    \item[] Justification: Appendix~\ref{app:ethics_compute} provides a summary table of the main external assets used in this work, including datasets, foundation models, APIs, and segmentation/localization tools. The original creators are cited in the main paper, and the corresponding licenses, provider terms, or original asset terms are specified and respected.

\item {\bf New assets}
    \item[] Question: Are new assets introduced in the paper well documented and is the documentation provided alongside the assets?
    \item[] Answer: \answerYes{} 
    \item[] Justification: The paper documents data sources, filtering, annotation, DINOv3 K-Means sampling, triplet construction, exact splits, and CC BY-NC 4.0 licensing. Appendix~\ref{app:data_balancing_annotation} distinguishes caption labels from sampling clusters and points to the README for complete count tables and taxonomy comparisons. The public dataset card documents fields and splits.

\item {\bf Crowdsourcing and research with human subjects}
    \item[] Question: For crowdsourcing experiments and research with human subjects, does the paper include the full text of instructions given to participants and screenshots, if applicable, as well as details about compensation (if any)? 
    \item[] Answer: \answerNo{}
    \item[] Justification: Appendix~\ref{app:ethics_compute} summarizes the evaluation procedure, rating dimensions, blinding, voluntary participation, and compensation policy. The full original participant instructions and interface screenshots are not included.

\item {\bf Institutional review board (IRB) approvals or equivalent for research with human subjects}
    \item[] Question: Does the paper describe potential risks incurred by study participants, whether such risks were disclosed to the subjects, and whether Institutional Review Board (IRB) approvals (or an equivalent approval/review based on the requirements of your country or institution) were obtained?
    \item[] Answer: \answerNo{} 
    \item[] Justification: The human perceptual study involved anonymous image-quality ratings and did not collect sensitive personal information from participants. Although no formal IRB approval or equivalent review was obtained, Appendix~\ref{app:ethics_compute} describes the minimal-risk nature of the study, informed participation, voluntary withdrawal, fair compensation, and anonymized aggregation of responses.

\item {\bf Declaration of LLM usage}
    \item[] Question: Does the paper describe the usage of LLMs if it is an important, original, or non-standard component of the core methods in this research? Note that if the LLM is used only for writing, editing, or formatting purposes and does \emph{not} impact the core methodology, scientific rigor, or originality of the research, declaration is not required.
    \item[] Answer: \answerYes{} 
    \item[] Justification: The paper explicitly uses Gemini 2.0 Flash for dense garment captioning and Qwen-VL-Plus as a VLM-based judge for semantic evaluation. These VLMs are important components of the dataset construction and evaluation protocol, and their usage is described in Sections 3 and 4 as well as the supplementary material.

\end{enumerate}
\end{document}